%% file: main.tex
\documentclass[sn-apa,iicol]{sn-jnl}

\usepackage{graphicx}%
\usepackage{multirow}%
\usepackage{amsthm}%
\usepackage{mathrsfs}%
\usepackage[title]{appendix}%
\usepackage{xcolor}%
\usepackage{textcomp}%
\usepackage{booktabs}%
\usepackage{listings}%
\usepackage{array}
\usepackage[caption=false,font=normalsize,labelfont=sf,textfont=sf]{subfig}
\usepackage{amsmath,amsfonts}
\usepackage{stfloats}
\usepackage{url}
\usepackage{verbatim}
\usepackage{bm}
\usepackage{bbm}
\usepackage{xspace}
\usepackage{enumitem}
\usepackage{pifont}
\usepackage{cleveref}
\usepackage{color}
\usepackage{colortbl}
\def \ie {\emph{i.e.}}
\def \eg {\emph{e.g. }}

\def \argmax {\text{argmax}}

\definecolor{Gray}{gray}{0.9}
\definecolor{LightCyan}{rgb}{0.88,0.95,1}
\newcommand{\tit}[1]{\smallbreak\noindent\textbf{#1.}}
\newcommand{\tinytit}[1]{\noindent\textbf{#1.}}

\newcommand{\ours}{PAC-S++\xspace}
\newcommand{\oursref}{RefPAC-S++\xspace}

\newcommand{\rev}[1]{\textcolor{black}{#1}}

\theoremstyle{thmstyleone}%
\theoremstyle{thmstyletwo}%

\theoremstyle{thmstylethree}%

\begin{document}
\title[Article Title]{Positive-Augmented Contrastive Learning for Vision-and-Language Evaluation and Training}

\author*[1]{\fnm{Sara} \sur{Sarto}}\email{sara.sarto@unimore.it}
\equalcont{\small{These authors contributed equally to this work.}}

\author*[1]{\fnm{Nicholas} \sur{Moratelli}}\email{nicholas.moratelli@unimore.it}
\equalcont{\small{These authors contributed equally to this work.}}

\author[1]{\fnm{Marcella} \sur{Cornia}}\email{marcella.cornia@unimore.it}

\author[1]{\fnm{Lorenzo} \sur{Baraldi}}\email{lorenzo.baraldi@unimore.it}

\author[1,2]{\fnm{Rita} \sur{Cucchiara}}\email{rita.cucchiara@unimore.it}

\affil[1]{\orgdiv{University of Modena and Reggio Emilia}, \orgaddress{\city{Modena}, \postcode{}\country{Italy}}}

\affil[2]{\orgdiv{IIT-CNR}, \orgaddress{\city{Pisa},\postcode{} \country{Italy}}}

\definecolor{TitleColor}{gray}{0.95}


\abstract{
Despite significant advancements in caption generation, existing evaluation metrics often fail to capture the full quality or fine-grained details of captions. This is mainly due to their reliance on non-specific human-written references or noisy pre-training data. Still, finding an effective metric is crucial not only for captions evaluation but also for the generation phase. Metrics can indeed play a key role in the fine-tuning stage of captioning models, ultimately enhancing the quality of the generated captions. In this paper, we propose \ours, a learnable metric that leverages the CLIP model, pre-trained on both web-collected and cleaned data and regularized through additional pairs of generated visual and textual positive samples. Exploiting this stronger and curated pre-training, we also apply \ours as a reward in the Self-Critical Sequence Training (SCST) stage typically employed to fine-tune captioning models. Extensive experiments on different image and video datasets highlight the effectiveness of \ours compared to popular metrics for the task, including its sensitivity to object hallucinations. Furthermore, we show that integrating \ours into the fine-tuning stage of a captioning model results in semantically richer captions with fewer repetitions and grammatical errors. Evaluations on out-of-domain benchmarks further demonstrate the efficacy of our fine-tuning approach in enhancing model capabilities. Source code and trained models are publicly available at: \url{https://github.com/aimagelab/pacscore}.
}




\maketitle

\input{sections/01_intro}
\input{sections/02_related}

\input{sections/03_metric}
\input{sections/04_reinforcement}

\input{sections/05_experiments}
\input{sections/06_conclusion}





\backmatter
\bmhead{Acknowledgements}
We acknowledge the CINECA award under the ISCRA initiative, for the availability of high-performance computing resources and support. This work has been conducted under two research grants, one co-funded by Leonardo S.p.A. and the other co-funded by Altilia s.r.l., and supported by the PNRRM4C2 project ``FAIR - Future Artificial Intelligence Research'' and by the PRIN 2022-PNRR project ``MUCES'' (CUP E53D23016290001), both funded by the European Union - NextGenerationEU.

\bmhead{Data availability}
Data sharing not applicable to this article as no datasets were generated during the current study. Datasets employed for this article are all publicly available.

\bibliography{bibliography}

\input{sections/suppl}

\end{document}

%% file: sections/01_intro.tex
\section{Introduction}\label{sec1}
The objective of image captioning is to provide natural language descriptions, conditioned on input images, that closely resemble human language and align with human intentions. This field has gained significant attention in recent years, resulting in captioning models capable of accurately describing images in detail. These advancements are due to methodological and architectural innovations~\citep{stefanini2022show}, as well as the use of larger pre-training datasets.

The evolution from early models based on recurrent neural networks~\citep{karpathy2015deep,xu2015show} to self-attentive architectures~\citep{huang2019attention,cornia2020meshed,pan2020x} represents significant advancements in image captioning research. These improvements have focused on better connecting visual and textual modalities and incorporating objects and tags at the architectural level~\citep{li2020oscar,zhang2021vinvl}. 
Additionally, there has been a notable emphasis on enhancing the robustness of cross-modal features~\citep{li2022comprehending,barraco2023little}, leading to more accurate captions. Today, image captioning has been integrated into multimodal large language models~\citep{li2023blip,liu2023visual,dai2023instructblip,chen2024sharegpt4v}, which show a strong ability to generate detailed and complex descriptions among other tasks.

As the quality of caption generation improves, developing automated methods for evaluating captions becomes even more crucial. The evaluation of captioning models should consider their ability to accurately describe images without hallucinations and closely align with human judgment. Moreover, an effective captioning metric should evaluate the content and style of generated captions, regardless of the significant variety of features that an image description might have. In some cases, to enhance the evaluation process, these metrics can also include comparisons to reference human-written captions.
Early attempts at captioning evaluation drew upon metrics born for machine translation~\citep{papineni2002bleu,lin2004rouge,banerjee2005meteor} or text-only domains~\citep{vedantam2015cider,spice2016,zhang2019bertscore}. However, these metrics often struggle to capture aspects such as grammatical correctness, semantic relevance, and specificity due to the different application domains. 
Moreover, despite their reliance on reference captions, these metrics sometimes penalize accurately generated captions that describe novel elements not covered in the reference sentences, thus leading to inaccuracies in evaluation.

Captioning metrics and reference captions are not only used for evaluation: some captioning models exploit them also to enhance their performance during generation. For instance, by optimizing a non-differentiable metric, such as CIDEr~\citep{vedantam2015cider}, captioning models can improve performance in a fine-tuning stage based on reinforcement learning after standard training with cross-entropy loss. This additional training stage that exploits the CIDEr metric as reward, known as Self-Critical Sequence Training (SCST)~\citep{rennie2017self}, has been widely adopted and can be considered as a \textit{de facto} standard in image captioning literature~\citep{stefanini2022show}.

\begin{figure}[t]
\begin{center}
    \begin{minipage}{0.98\linewidth}
    \includegraphics[width=0.98\linewidth]{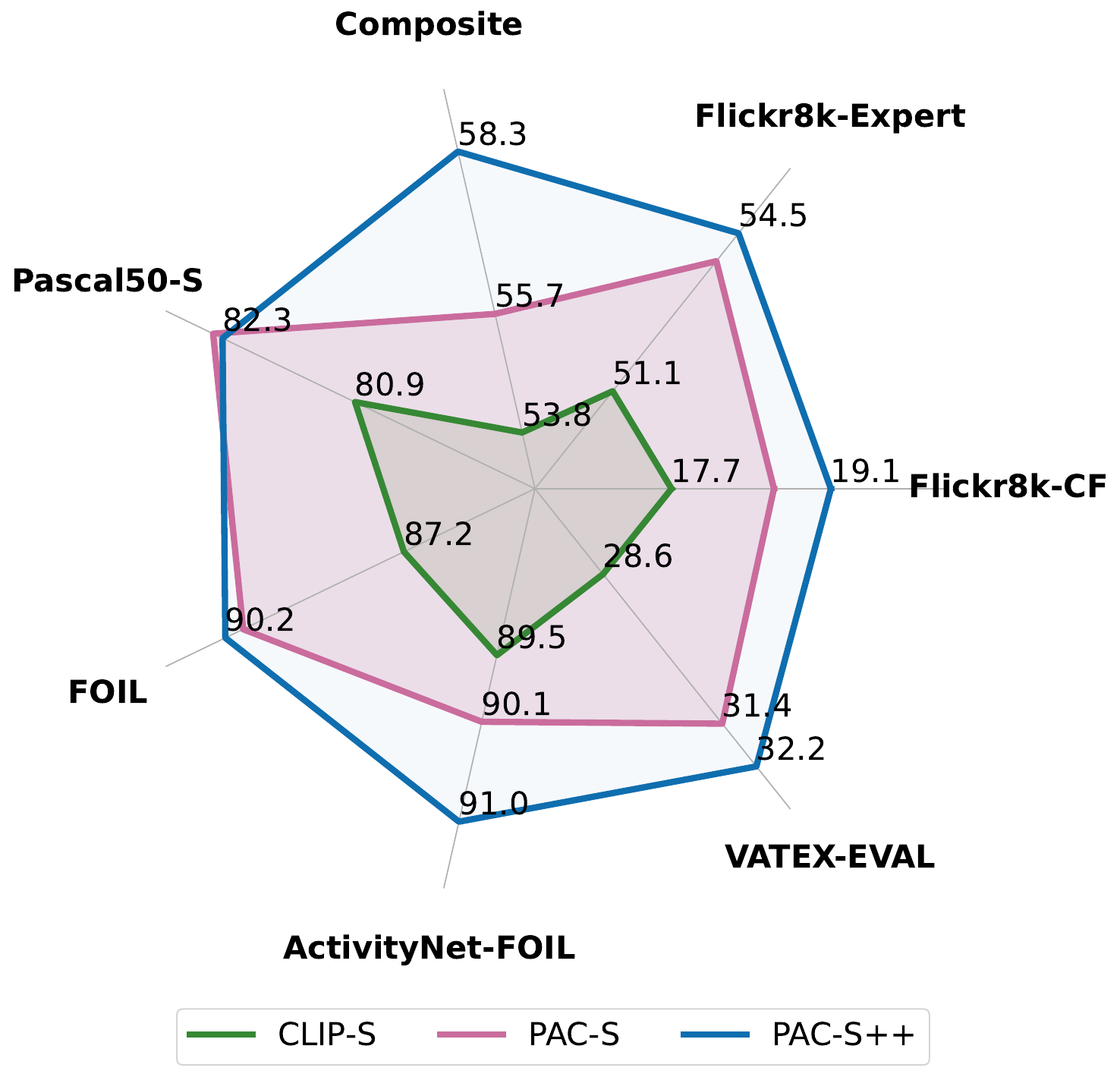}
    \end{minipage}
    \vspace{0.15cm}
    \begin{minipage}{\linewidth}
    \includegraphics[width=\linewidth]{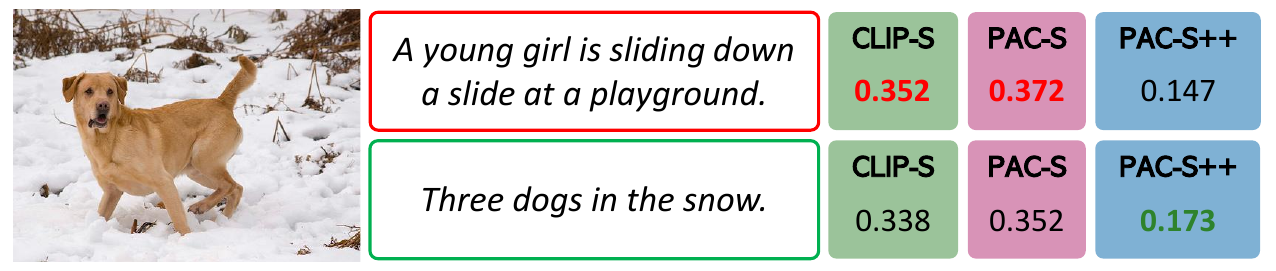}
    \end{minipage}
    \vspace{-0.2cm}
\end{center}
\caption{Comparison between evaluation scores predicted by our evaluation metric, \ours, in comparison with its original version, PAC-S~\citep{sarto2023positive}, and CLIP-S~\citep{hessel2021clipscore}. The plot shows the results across different benchmarks, demonstrating the superior performance of \ours in terms of correlation with human judgment. In the bottom example, the caption highlighted in green is the one preferred by humans.}
\label{fig:radarplot}
\end{figure}

To enhance alignment with human judgment and address the limitations of standard captioning metrics (\eg~grammatical and semantic correctness, specificity, etc.), a set of advanced metrics that align visual and textual data have recently emerged~\citep{hessel2021clipscore,shi2022emscore,wada2024polos}. A notable trend in these metrics is to leverage the multimodal CLIP embedding space~\citep{radford2021learning} that, when exploited in evaluation, exhibits improved correlation with human judgment, especially thanks to the larger scale of the underlying architecture and the amount of pre-training data. However, to sustain such an increase in the amount of training data, large-scale multimodal models like CLIP usually exploit image-text pairs crawled from the web~\citep{sharma2018conceptual,schuhmann2022laion}, resulting in noisy collections whose style and distribution are not aligned with those on which captioning systems are evaluated~\citep{lin2014microsoft}. Clearly, this lack of data quality can potentially limit the effectiveness of captioning metrics that are developed on top of the resulting embedding spaces.

The ideal solution to the aforementioned issue would be training on cleaned data sources, which are however limited in size. As an alternative, we propose a learnable metric that incorporates the richness of pre-training on web-collected data as well as the quality of cleaned data. To sustain the need for quantity, we regularize the training of the CLIP embedding space by including additional positive samples generated from both visual~\citep{ramesh2022hierarchical,rombach2022high} and textual~\citep{zhang2021vinvl,li2022blip} generators. These generators enable the synthetic generation of data in both modalities, allowing for controlled style and quality. Our proposed metric, termed as \ours, is trained via a novel positive-augmented contrastive learning approach, in which pairs of generated images and texts act as supplementary positives in addition to real images and human-annotated captions taken from a cleaned data source. To regularize training, we employ low-rank adaptation~\citep{hu2021lora} that can enhance the final performance while preserving the original advantages of the CLIP embedding space.

Since captioning metrics should be able to judge the alignment between image-caption pairs, beyond the standard cross-entropy loss employed to train captioning models, they can also serve as a positive signal to enhance the semantic richness and descriptiveness of generated captions. In addition to the use of the standard CIDEr metric for fine-tuning captioning models, metrics like CLIP-S have been employed as well in the SCST fine-tuning stage~\citep{cho2022fine}, where they are utilized as reward signals. Despite some improvements in the richness of the final descriptions, these solutions often lead to excessively long and repetitive captions. To address this, we propose to employ \ours as reward for fine-tuning captioning models, leveraging the fact that our metric does not rely on human references by design and is based on an improved image-text alignment, unlike CIDEr and CLIP-S respectively. 

To evaluate the effectiveness of our metric, we conduct extensive experiments across diverse datasets and settings with the aim of assessing the correlation degree with human judgment and determining whether it can be effectively employed as reward signal during the fine-tuning stage of captioning models. Specifically, datasets like Flickr8k-Expert and Flickr8k-CF~\citep{hodosh2013framing}, Composite~\citep{aditya2015images}, and Pascal-50S~\citep{vedantam2015cider} are employed to evaluate the correlation of image-caption pairs, while the VATEX-EVAL dataset~\citep{shi2022emscore} is used for the video scenario. Further, we assess the sensitivity of the proposed metric to object hallucination, performing experiments on the FOIL~\citep{shekhar2017foil} and ActivityNet-FOIL~\citep{shi2022emscore} dataset (Fig.~\ref{fig:radarplot}). Finally, by conducting experiments on standard captioning benchmarks such as COCO~\citep{lin2014microsoft}, nocaps~\citep{agrawal2019nocaps}, VizWiz~\citep{gurari2020captioning}, and CC3M~\citep{sharma2018conceptual}, we demonstrate that training a captioning model using \ours as reward can lead to semantically richer image descriptions, while not compromising their grammatical correctness.

In summary, our proposed metric outperforms previous reference-based and reference-free evaluation scores, demonstrating superior performance compared to CLIP-S~\citep{hessel2021clipscore} and the corresponding video-based version (\ie~EMScore~\citep{shi2022emscore}), which also employ a contrastive embedding space for evaluating image/video-caption pairs. Moreover, when employed as a reward in the SCST fine-tuning stage, \ours leads to richer captions with fewer hallucinations and grammatical errors. 

This work is an enhanced and extended version of our conference paper~\citep{sarto2023positive}. In contrast to our prior work, the proposed evaluation metric is extended by introducing a low-rank fine-tuning stage which preserves the pre-trained model weights while injecting trainable rank decomposition matrices. Moreover, we introduce \ours as a reward during the SCST phase to improve captioning models, resulting in captions with enriched semantics.

%% file: sections/02_related.tex
\section{Related Work}
\tit{Standard Metrics}
Captioning evaluation aims to assess the quality of a generated caption describing a given image or video, optionally based on human-annotated reference captions. Many widely used captioning evaluation metrics were originally developed in the context of NLP tasks and rely on $n$-gram matching techniques. These classical metrics include 
BLEU~\citep{papineni2002bleu}, METEOR~\citep{banerjee2005meteor}, and ROUGE~\citep{lin2004rouge}. Specifically, BLEU and METEOR were introduced for machine translation. BLEU relies on $n$-gram precision, while METEOR prioritizes the recall of matching unigrams between candidate and reference sentences, considering their exact form, stemmed form, and semantic meaning. ROUGE, instead, was designed for summarization tasks and adapted for evaluating image or video descriptions.

Later, two metrics tailored for the captioning task emerged, namely CIDEr~\citep{vedantam2015cider} and SPICE~\citep{spice2016}. The former assesses $n$-gram cosine similarity based on TF-IDF (Term Frequency-Inverse Document Frequency) taking into account both precision and recall, and the latter quantifies graph-based similarity using scene graphs derived from candidate and reference captions. These metrics primarily concentrate on textual-level comparisons, operating under the assumption that the information conveyed in human-written references accurately represents the image content.

\tit{Learning-based Metrics}
While traditional metrics are primarily based on text alignment between reference and machine-generated captions, several captioning metrics that also take the visual input into account have been developed in recent years. Some of them, such as TIGER~\citep{jiang2019tiger}, consider word-image region similarities to compute the final score. With the introduction of large pre-trained models, however, the most common trend involves exploiting the capabilities of these architectures to evaluate the coherence of a given caption with the input image or video and eventually reference sentences~\citep{lee2020vilbertscore,wang2021faier,lee2021umic}.

In this context, the CLIP model~\citep{radford2021learning} is the most widely used large-scale multimodal model for the task, with the CLIP-Score~\citep{hessel2021clipscore} being the first metric based on a modified cosine similarity between image and candidate caption representations extracted from CLIP visual and textual encoders. Following this line of research, MID~\citep{kim2022mutual} uses CLIP visual-textual features to compute negative Gaussian cross-mutual information, resulting in a more effective evaluation metric. Parallel efforts have been made in the evaluation of video descriptions, exemplified by the EMScore~\cite{shi2022emscore}, which computes fine-grained similarities between video frames and words of the candidate caption using CLIP embeddings.  
More recent metrics still utilize multimodal models (\ie~CLIP) but incorporate additional components for enhanced performance. For instance, BRIDGE~\citep{sarto2024bridge} employs a mapping module to generate pseudo-captions that capture more fine-grained visual details. Similarly, HICE-S~\citep{zeng2024hicescore} introduces a hierarchical scoring mechanism that identifies local visual regions and textual phrases using the Segment Anything Model (SAM)~\citep{kirillov2023segment}. In contrast, Polos~\citep{wada2024polos} is a supervised evaluation metric that fine-tunes the CLIP embedding space on a dedicated dataset.

On a different line, some solutions exploit the effectiveness of language models to evaluate generated sentences, initially comparing them with ground-truth captions using BERT-based embeddings~\citep{zhang2019bertscore,yi2020improving} and then leveraging the extensive pre-training and capabilities of large language models, like GPT-3.5, to obtain more effective evaluation scores~\citep{chan2023clair,lee2024fleur}.

Another crucial challenge in evaluating generated captions is detecting the presence of errors, such as the hallucination of objects that are not present in the image. Recent studies delve into addressing the well-known problem of hallucination, such as the CHAIR~\citep{rohrbach2018object} and ALOHa~\citep{petryk2024aloha} metrics.

\tit{Image Captioning and Training Strategies}
Aligning models with human judgment remains a significant challenge not only in evaluation but also in generation. Early models, ranging from CNN-based encoders and RNNs~\citep{vinyals2015show,karpathy2015deep} to the latest fully attentive architectures~\citep{huang2019attention,pan2020x,cornia2020meshed,li2022comprehending}, generate captions by greedily selecting the most probable word from a learned vocabulary. To mitigate error propagation during generation, the beam search algorithm~\citep{koehn2009statistical} has become widely adopted. This algorithm maintains a set of $k$ most likely sequence candidates and ultimately outputs the most probable sequence from this set.

Captioning models learn probability distributions that mirror human-annotated sentences. Most approaches utilize a combination of cross-entropy loss for pre-training and reinforcement learning strategies, such as the Self-Critical Sequence Training (SCST)~\citep{rennie2017self}, for fine-tuning.
While the cross-entropy loss minimizes the negative log-likelihood of ground-truth tokens, SCST maximizes the expected reward by comparing generated and ground-truth captions employing a non-differentiable evaluation metric (\ie~usually the CIDEr score). This approach yields more accurate and human-like descriptions compared to cross-entropy alone. Consequently, this training paradigm has become a standard~\citep{stefanini2022show}. 

However, the emergence of pre-trained vision-and-language models like CLIP~\cite{radford2021learning} has highlighted the limitations of traditional metrics for evaluating caption quality and, consequently, for using them as a reward during SCST. In fact, while the use of a CIDEr-based reward can help align generated captions with ground-truth examples, it often reduces the semantic richness of the predicted sentences. To solve this issue, there has been limited exploration of learnable reward models that align references and generated captions without handcrafted metrics~\citep{cho2022fine,dessi2023cross}. Additionally, there is a growing interest in exploiting the large-scale pre-training of large language models to obtain semantically richer descriptions. In this context, some approaches~\citep{ramos2023smallcap,mokady2021clipcap} employ pre-trained language models like GPT-2 and exclusively train 
specific components, such as cross-attention layers, to capture the complex relationships between images and corresponding textual descriptions. Other solutions~\citep{rotstein2024fusecap,dong2024benchmarking}, instead, directly adapt multimodal large language models to generate more detailed captions.

%% file: sections/03_metric.tex
\section{PAC-Score++}
We aim to develop an image and video captioning metric based on a shared embedding space where visual data and text can be represented and evaluated. To achieve this, we adopt the dual-encoder architecture introduced by CLIP~\citep{radford2021learning}, enhancing it through fine-tuning with low-rank adaptation (LoRA) techniques~\citep{hu2021lora}. We show that our metric can also be applied for the fine-tuning stage of captioning models to improve the quality and descriptiveness of generated captions.

\begin{figure*}[t]
    \centering
    \includegraphics[width=0.99\linewidth]{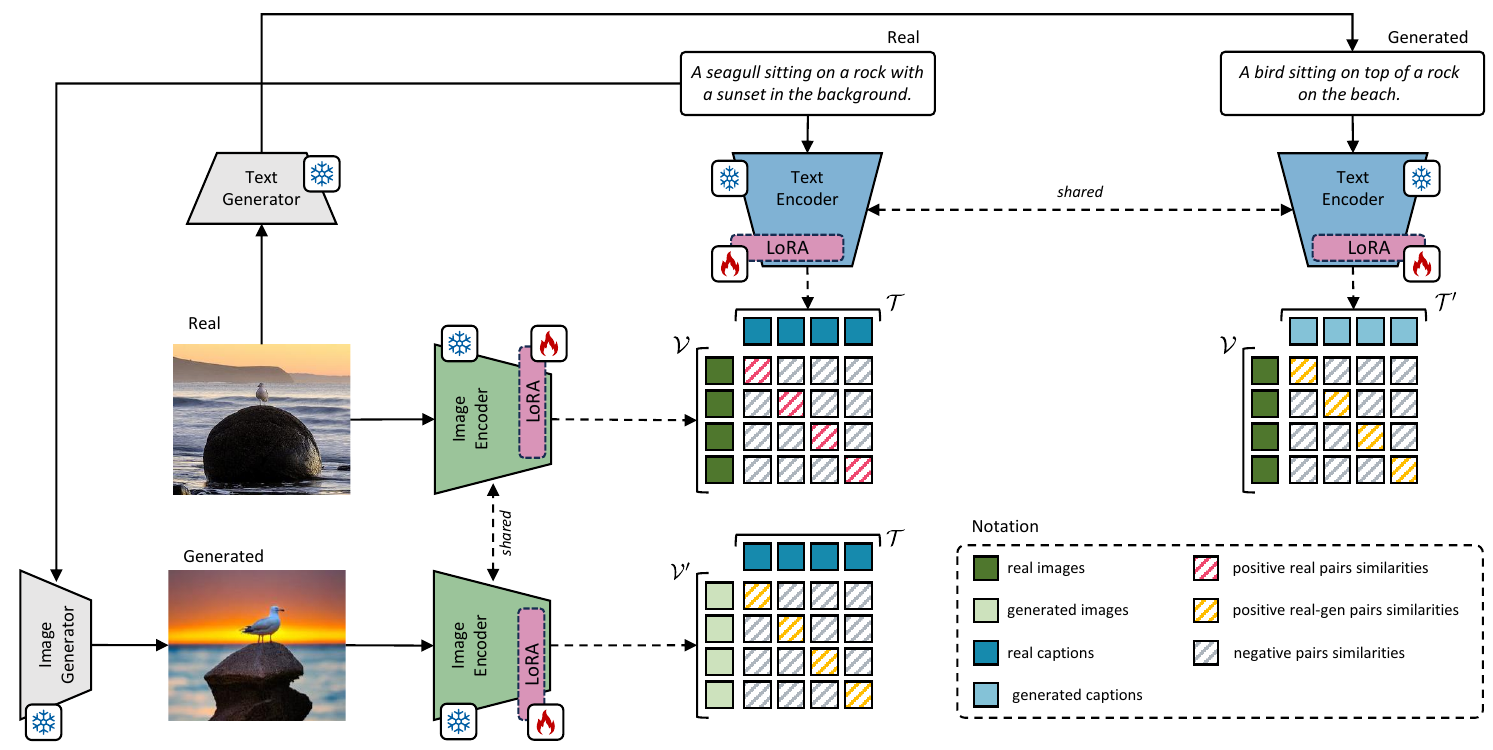}
    \caption{Overview of our positive-augmented contrastive learning approach in which both encoders are fine-tuned with low-rank adaptation (LoRA) using additional positive samples generated by text-to-image and image-to-text generative models.}
    \vspace{-0.1cm}
    \label{fig:model}
\end{figure*}

\subsection{Revisiting CLIP}
Contrastive Language-Image Pre-training (CLIP) focuses on learning rich visual and textual representations by understanding the relationships between images and their corresponding textual descriptions. CLIP employs an image encoder $E_v(\cdot)$ (\eg~a CNN~\citep{he2016deep} or a ViT~\citep{dosovitskiy2020image}) along with a text encoder $E_t(\cdot)$ (\eg~a Transformer model~\citep{vaswani2017attention}) to obtain visual and textual representations. 
The multimodal interaction is performed via late fusion by projecting the output of both encoders to the same dimension and then on the $\ell_2$ hypersphere via normalization. The visual and the textual inputs can then be compared via cosine similarity.

During the training phase, CLIP utilizes a contrastive objective to encourage similar embeddings for matched image-text pairs and dissimilar embeddings for non-matched pairs. In a batch of $N$ image-caption pairs ${\{(v_i, t_i)\}}_{i=1}^{N}$, CLIP employs the InfoNCE loss~\citep{oord2018representation} that can be written as:
\begin{gather}
\label{eq:infonce}
    \mathcal{L}_{\mathcal{V}, \mathcal{T}} = -\frac{1}{N} \sum_{i=1}^N \log \frac{\exp(\text{sim}(v_i, t_i) / \tau)}{\sum_{j=1}^N \exp(\text{sim}(v_i, t_j) / \tau)} + \\
    \nonumber
    -\frac{1}{N} \sum_{i=1}^N \log \frac{\exp(\text{sim}(v_i, t_i) / \tau)}{\sum_{j=1}^N \exp(\text{sim}(v_j, t_i) / \tau)}.
\end{gather}
Here, the similarity function is defined as:
\begin{equation}
\nonumber
    \operatorname{sim}(v, t) = \cos(\text{Norm}(E_v(v)), \text{Norm}(E_t(t))),
\end{equation}
where $\text{sim}(\cdot)$ is the CLIP-based cosine similarity between visual and textual inputs that are normalized via $\ell_2$ normalization, and $\tau$ is a temperature parameter to scale the logits. With the symmetrical loss applied to both image and text encoders, the overall loss function $L_{\mathcal{V}, \mathcal{T}}$ is computed as the average of the two.

Large-scale contrastive models like CLIP are trained using image-caption pairs collected from the web. These provide a large-scale source of supervision for learning scalable low-level and semantic visual and textual features, as testified by their zero-shot classification performance and by their adaptability to different tasks~\citep{ramesh2022hierarchical,materzynska2022disentangling,khandelwal2022simple}. However, it should be noted that the textual annotations contained in alt-tags are not of the same quality expected by evaluators. Additionally, the distribution of images at the web-scale may not be perfectly aligned with those used to evaluate image captioning systems.

To address this concern, an intuitive solution could involve training the embedding space directly on cleaned data sources. However, recent attempts to learn contrastive-based evaluation metrics on curated datasets like COCO~\citep{lin2014microsoft} have shown poor performance compared to traditional metrics, potentially because of the lack of training data~\citep{jiang2019tiger}.

\subsection{Positive-Augmented Contrastive Learning}\label{subsect:pac_method}
In light of these problems, we propose utilizing synthetic generators for both visual and textual data, which showcase sufficiently high-quality levels of generation. Additionally, they are controllable in terms of visual distribution. 

Specifically, given a positive image-text pair $(v, t)$, we augment it by generating a synthetic caption $t'$ from $v$ using an image captioning model~\citep{li2022blip}. Similarly, we generate a synthetic image $v'$ from $t$ via a diffusion-based text-to-image architecture~\citep{rombach2022high}, thus building a dataset consisting of tuples of four elements $(v, t, v', t')$.
Next, we train our evaluation model by considering the contrastive relationships between real and generated matching image-caption pairs, as shown in Fig.~\ref{fig:model}.
By introducing low-rank decompositions into the network parameters, we obtain a fine-tuned visual encoder $E_v(\cdot)$ and text encoder $E_t(\cdot)$. Specifically, we employ LoRA~\citep{hu2021lora} which preserves the pre-trained model weights while injecting trainable rank decomposition matrices into each layer of the architecture. This approach significantly reduces the overall number of trainable parameters, mitigates the risk of overfitting, and regularizes the training procedure, thus making it a suitable option for the fine-tuning phase.

Formally, given a batch of $N$ real images and their captions, these are processed through the corresponding encoders to obtain the visual $\mathcal{V} = \left\{E_v(v_{i})\right\}_{i=1}^{N}$ and textual features  $\mathcal{T} = \left\{E_t(t_{i})\right\}_{i=1}^{N}$. For generated images and texts, we define $\mathcal{V}' = \left\{E_v(v'_{i})\right\}_{i=1}^{N}$ and 
$\mathcal{T}' = \left\{E_v(t'_{i})\right\}_{i=1}^{N}$. We then define multiple $N \times N$ matrices containing pairwise cosine similarities between the different inputs. We then adopt a symmetric InfoNCE loss, which aims at maximizing the cosine similarity between the $N$ matching pairs and minimizing those of the $N^2 - N$ non-matching pairs. 

In addition to the loss term between real images and real texts $\mathcal{L}_{\mathcal{V}, \mathcal{T}}$, defined in Eq.~\ref{eq:infonce}, we also add symmetrical loss terms between cross-modal generated and real pairs, \ie~between generated images and human-annotated texts, and between original images and generated texts. The loss which compares real images $\mathcal{V}$ with respect to generated texts $\mathcal{T}'$ can be defined as:
\begin{gather}
\label{eq:loss_real_v_gen_t}
\nonumber
    \mathcal{L}_{\mathcal{V}, \mathcal{T}'} = -\frac{1}{N} \sum_{i=1}^N \log \frac{\exp(\text{sim}(v_i, t_i') / \tau)}{\sum_{j=1}^N \exp(\text{sim}(v_i, t_j')) / \tau)} + \\
    -\frac{1}{N} \sum_{i=1}^N \log \frac{\exp(\text{sim}(v_i, t_i')/ \tau)}{\sum_{j=1}^N \exp(\text{sim}(v_j, t_i') / \tau)}.
\end{gather}
In this way, generated items act as additional positive samples for the real matching pairs, thus adding a supervisory signal without being affected by the potential noise present in the data used to train contrastive-based feature extractors like CLIP. In summary, the final loss is a weighted combination of the three loss terms, \ie 
\begin{equation}
\label{eq:final_loss}
    \mathcal{L} = \mathcal{L}_{\mathcal{V}, \mathcal{T}} + \lambda_v \mathcal{L}_{\mathcal{V}', \mathcal{T}} + \lambda_t  \mathcal{L}_{\mathcal{V}, \mathcal{T}'},
\end{equation}
where $\mathcal{L}_{\mathcal{V}', \mathcal{T}}$ is the counterpart of Eq.~\ref{eq:loss_real_v_gen_t} using generated image and real textual sentences, and the $\lambda$ values are hyperparameters used to weight the contribution of each loss function.

\subsection{Evaluating Image-Caption Pairs}
Starting from the trained embedding space with positive-augmented contrastive learning, an evaluation metric for image captioning can be defined by simply scaling, and eventually thresholding, the similarity computed inside of the embedding space itself. For evaluating images, we adopt the equation proposed by~\cite{hessel2021clipscore} as our reference-free score:
\begin{equation}
    \label{eq:clip_score}
    \text{Score}(v, t) = w \cdot \max (\text{sim}(v, t), 0),
\end{equation}
that given an image-text pair ${(v,t)}$ defines the evaluation score as a linear projection of thresholded cosine similarities.

To incorporate reference ground-truth captions into the evaluation process, following~\citep{hessel2021clipscore}, we first calculate the representation of each reference caption using our positive-augmented trained textual encoder. Then, we compute the harmonic mean between the reference-free score, defined in Eq.~\ref{eq:clip_score}, and the maximum cosine similarity between the candidate caption and all reference captions. 
Formally, given a set of $M$ reference captions $R = \left\{r^j\right\}_{j=1}^{M}$, the score is computed as:
\begin{equation}
    \begin{aligned}
     \text{Ref-Score}(v, t, R) & =
         \text{H-Mean}\left(\text{Score}(v, t), \text{top-r}(t)\right) \\[.5em]
       \text{where} \quad \text{top-r}(t) & = \max\left(0, \max_{r \in R}(\text{sim}(t, r))\right). 
    \end{aligned} 
    \label{eq:refclips}
\end{equation}
Here, $\text{Score}(\cdot)$ represents the reference-free score defined in Eq.~\ref{eq:clip_score}, and $\text{H-Mean}(\cdot)$ indicates the harmonic mean.

\subsection{Evaluating Video-Caption Pairs}
To evaluate video captions using the positive-augmented strategy, we expand upon the previously defined metric following the approach proposed by~\cite{shi2022emscore}. Specifically, we use our trained embedding space to extract video and text embeddings at both fine-grained and coarse-grained levels.

Given a video $W=\{w^j\}_{j=1}^{|W|}$ , where $|W|$ is the number of frames, each fine-grained frame embedding $\mathcal{W}_{f}^{j}$ is obtained as follows:
\begin{equation}
		\mathcal{W}_{f}^{j} = \text{Norm}\left( E_{v}(w^j)\right),
\end{equation}
where Norm$(\cdot)$ is the $\ell_2$ normalization function.

The coarse-grained video embedding $\mathcal{W}_c$ is obtained by 
normalizing the mean-pooling of all frame embeddings:
\begin{equation}
	\begin{footnotesize}
		\mathcal{W}_c=\operatorname{Norm}\left(\frac{1}{|W|} \cdot \sum_{j=1}^{|W|} \mathcal{W}_{f}^{j} \right).
	\end{footnotesize}
\end{equation}

For a given caption $t$, the CLIP tokenizer, that adds two special tokens \texttt{[SOS]} and \texttt{[EOS]} respectively at the beginning and the end of the sentence, is used to construct a new token sequence of length $|L|$ which is then passed through the CLIP textual encoder. Formally, we define 
\begin{align}
        t_f & = W \cdot \text{LN}(\hat{E}_{t}(t)) \\
        \nonumber
         & = \{{t}_f^{\text{\texttt{SOS}}}, {t}_f^{{1}}, \cdots, {t}_f^{{|L|-2}},{t}_f^{\text{\texttt{EOS}}}\}, 
\end{align}
where $\hat{E_t}(\cdot)$ is the CLIP text encoder before the last layer normalization (LN) and linear projection $W$. 
Each of the $|L|$ token embeddings is used for fine-grained embedding matching, while the \texttt{[EOS]} token serves as the global embedding for coarse-grained embedding matching. Specifically, 
we define $E_t(t) = t_f^{\text{\texttt{EOS}}}$, which we denote as $t_c$ for the sake of notation. 

\tit{Coarse-grained Embedding}
Given the source video $W$ and the caption $t$, the coarse-grained score can be computed as the inner product between the corresponding coarse-grained embeddings: 
\begin{equation}
    \label{eq:ems_score_coarse}
    \text{Score}(W, t)_{c} = \mathcal{W}_c^{\top}t_c.
\end{equation}
This comparison evaluates the overall similarity between the video and the caption at a higher level, capturing the coarse-grained alignment between the two.

\tit{Fine-grained Embedding}
Relying solely on coarse-grained embedding matching may result in a loss of detailed information due to the changing visual elements in each frame. To address this, a fine-grained embedding matching approach is introduced to establish alignment between individual frames and sentence tokens, enabling a more detailed evaluation of video captions.

Given the video frame embedding $\mathcal{W}_f$ and the sentence token embedding ${t}_f$, precision $P(\cdot)$ and recall $R(\cdot)$ are computed. Specifically, the precision evaluates whether descriptions are related
to the video content without incorrect details. Moreover, to remove the visual-irrelevant words (\eg~``a'', ``the", ``and"), the inverse document frequency (IDF) is computed to model the importance of each word. After calculating the IDF values for the $l$-th word in the initial caption, the standard precision formulation is changed to:
\begin{equation}
	P(W, t)_{f}= \frac{\sum_{l=0}^{L-1}\operatorname{IDF}_l \cdot \max_{
   j} \left({\mathcal{W}_f^j}^{\top}t_f^l\right)}{\sum_{l=0}^{L-1} \operatorname{IDF}_l}.
\end{equation}
On the other hand, the recall computes the comprehensiveness of the caption, such as whether the content of the video is described without omission. Formally, the recall can be written as:
\begin{equation}
	R(W, t)_{f} = \frac{1}{|W|} \sum_{{j}} \max _{l}  \left({\mathcal{W}_f^j}^{\top}{t^l_f}\right).
\end{equation}

Finally, the fine-grained embedding score is defined as the F1 score that combines the evaluation of both recall and precision:
 \begin{equation}
	\label{equation: fine_F}
        \text{Score}(W, t)_{f} = 2 \cdot \frac{P(W,t)_f \cdot R(W,t)_f}{P(W, t)_f+R(W,t)_f}.
\end{equation} 

\tit{Final Evaluation Score}
The formulation for a reference-free setting for evaluating video-caption pairs is the average between the coarse and fine-grained scores:
\begin{equation}
		\label{equation: EMScore(X, W)}
		\begin{aligned}
			\text{Score}(W, t) = \frac{\text{Score}(W, t)_{c} + \text{Score}(W, t)_{f}}{ 2 } .
		\end{aligned}
\end{equation}

Also in this setting, we can integrate the reference caption $t_R$, if available, to compute a reference-based score, which is defined as the average of $\text{Score}(W, t)$ and $\text{Score}(t, t_R)$:
\begin{multline}
    \text{Ref-Score}(W, t, t_R) =\\ \frac{\text{Score}(W, t) + \text{Score}(t, t_R) }{ 2 },
    \label{equation: EMScore(X, V, X*)}
\end{multline}

where $\text{Score}(t, t_R)$ is computed following Eq.~\eqref{equation: EMScore(X, W)}, replacing the video $W$ with $t_R$.
If there are multiple $M$ reference sentences $\{t_R^{i}\}_{i=1}^{M}$, the reference-based score can still be computed by taking the maximum score between the target sentence and each reference sentence. 

%% file: sections/04_reinforcement.tex
\section{Fine-tuning Captioning Models with \ours}
\label{sec:SCST}
Leveraging reinforcement learning to optimize captioning metrics has become a widespread strategy to optimize image captioning systems and entails conceptualizing models as agents, with the primary goal of maximizing the expected reward. Inspired by the use of CIDEr and similar metrics, we explore the use of our metric, \ours, as a reward for fine-tuning a captioning model.

\subsection{Revisiting Standard Self-Critical Sequence Training}
Self-Critical Sequence Training (SCST)~\citep{rennie2017self} for image captioning is a two-step training methodology which (i) pre-trains a captioning network $f_\theta$ using a time-wise cross-entropy loss, and (ii) fine-tunes the same network by maximizing the CIDEr score~\citep{vedantam2015cider} on the training set using reinforcement learning.

While SCST effectively improves the quality of generated captions over single-stage cross-entropy training, it has been shown to introduce a bias towards generating captions that conform to the ``average'' description of the training set~\citep{chen2022learning}. This results in less descriptive, semantically rich, and discriminative captions. Moreover, these problems are amplified by uninformative image-caption pairs in captioning datasets, and by the reliance on the CIDEr metric as a reward signal, which has been questioned due to its relatively low correlation with human judgments and dependence on reference captions.

Recent attempts to replace CIDEr with semantic embedding-based metrics, like CLIP-S~\citep{cho2022fine}, have led to excessively long captions that, while detailed, may contain errors, \eg~repetitions, due to the noisy nature of the large-scale data used for CLIP pre-training.

\subsection{Self-Critical Sequence Training with \ours}
By combining pre-training on both web-collected and cleaned data, our metric addresses many of the issues associated with CIDEr and CLIP-S. 
As demonstrated in our previous work~\citep{sarto2023positive}, this approach results in a more refined embedding space and stronger correlations with human judgments. Consequently, we propose using \ours to improve the training of image captioning models.

\tit{First Training Stage (Cross-Entropy Loss)}
Formally, we can assume that $f_\theta$ is an autoregressive Transformer-based captioning network~\citep{vaswani2017attention}, where $\theta$ represents the trainable parameters, which takes as input an image $v$, described with a sequence of $R$ visual features 
$\{e^i\}_{i=1}^{R}$,
and a ground-truth sequence $t$ of words within the vocabulary. Notably, $\{e^i\}_{i=1}^{R}$
represents the grid of features before the last layer normalization and linear projection $W$ in the CLIP architecture:
\begin{equation}
    E_v(v) = W \cdot\text{LN}(e^1, \dots,e^R).
\end{equation}
During the first training stage, the network is conditioned on all visual features and all ground-truth tokens of length $T$ up to the current prediction step $k$. The model $f_\theta$ is optimized using the cross-entropy loss (\ie~\textit{teacher forcing}):
\begin{multline}
\label{eq:xe}
\mathcal{L}_{\text{XE}}(v, t ; \theta) =\\ - \sum_{k=1}^T \log f_\theta(t^k | t^1, \dots, t^{k-1}, e^1, \dots, e^R),
\end{multline}
where $f_\theta$ outputs a categorical probability distribution over the vocabulary.

\tit{Second Training Stage (SCST)}
In the second training stage, designed to enhance the generative capabilities of the model, the network is conditioned on the input image and previously generated words. The output of the captioning model $f_\theta$ is a generated caption $\hat{t} = \{\hat{t}^i\}_{i=1}^{S}$ of length $S$, where each word is sampled from the output probability distribution generated at the prior time step $k$.
For instance, the $k$-th token $\hat{t}^k$ is chosen as the one that maximizes the model probability distribution over possible tokens: 
\begin{equation}
    \hat{t}^k = \argmax f_{\theta}(\hat{t^k}|\hat{t}^{k-1}, ..., \hat{t}^1, e^1, ..., e^R).
\end{equation} 

Given the caption ${\hat{t}}$ and the image $v$, \ours score is computed and used as the reward $r(\cdot)$ for guiding a policy-gradient reinforcement learning update step: 
\begin{equation}
    r(v, \hat{t}) = \text{Score}(v, \hat{t} ),
\end{equation}
where $\text{Score}(\cdot)$ is computed as in Eq.~\ref{eq:clip_score}. Additionally, we consider a variant that takes into account reference captions, thus employing Eq.~\ref{eq:refclips} to compute the reward. To mitigate the variance in the reward signal, a baseline value $b$, computed as the average of the reward of all descriptions generated for $v$, is subtracted from the reward.

The parameters are optimized using gradient-based methods with the SCST loss function~\citep{rennie2017self}. Beam search is employed to explore multiple possible sequences. Formally, 
\begin{multline}
    \nabla_{\theta}\mathcal{L}_{\text{SCST}}(v, \hat{t}; \theta) = \\ -\frac{1}{l}\sum_{i=1}^{l}(r(v, \hat{t}^i) - b) \nabla_{\theta}\log f_\theta (\hat{t}^i)),
\end{multline}

where $l$ denotes the beam size and $t_i$ represents the $i$-th sentence in the beam.

%% file: sections/05_experiments.tex
\section{Experimental Evaluation}

\subsection{Implementation Details}

\tinytit{Positive-Augmented Contrastive Learning}
As commonly used in other CLIP-based evaluation metrics~\citep{hessel2021clipscore,kim2022mutual,shi2022emscore}, we employ CLIP ViT-B/32 as backbone to encode images or video frames and textual sentences. \rev{Moreover, we report some results using CLIP, SigLIP~\citep{zhai2023sigmoid} and SigLIP2~\citep{tschannen2025siglip2multilingualvisionlanguage} with the ViT-L/14 model to demonstrate the generalizability of our approach to more powerful backbones.} To refine the visual and textual representations of the model, we fine-tune visual and textual encoders using the methodology outlined in Sec.~\ref{subsect:pac_method}, utilizing the COCO dataset~\citep{lin2014microsoft} that consists of over 120,000 images accompanied by five captions. In particular, we employ the splits introduced by~\cite{karpathy2015deep}, where 5,000 images are used for validation, 5,000 images are used for testing, and the rest for training. 
During fine-tuning, we freeze the pre-trained model weights and exploit LoRA~\citep{hu2021lora}. The rank of the decomposition $r$ is set to 4, as it performed favorably in our initial experiments. We use AdamW~\citep{loshchilov2019decoupled} as optimizer with a learning rate equal to $1\cdot10^{-4}$ and a batch size of 256. The $\lambda_v$ and $\lambda_t$ values are selected with a grid search, choosing the combination that provides the best average across datasets. Specifically, we set $\lambda_v$ to 0.1 and $\lambda_t$ to 0.001, and stop the training stage when the validation loss stops decreasing for 1,500 iterations.

\tit{Positive Image-Text Generation} To expand the training dataset with additional positive instances, we use Stable Diffusion~\citep{rombach2022high} for generating new visual data and the BLIP model~\citep{li2022blip} for generating new textual descriptions. Specifically, to generate images, we employ the model pre-trained on the English image-text pairs of the LAION-5B dataset~\citep{schuhmann2022laion} and fine-tuned at a resolution equal to $512\times512$ on the LAION-Aesthetics subset\footnote{\url{https://laion.ai/blog/laion-aesthetics/}}, which has been filtered with aesthetic requirements. Throughout the generation process, we utilize a safety checker module to minimize the probability of explicit images. Moreover, we disable the invisible watermarking of the outputs to prevent easy identification of the images as being machine-generated.

\tit{Fine-tuning with RL} 
When assessing the effectiveness of \ours for fine-tuning captioning models, we employ a standard encoder-decoder Transformer architecture. Specifically, we use three layers in both encoder and decoder, a hidden size of 512, and 8 attention heads. To encode input images, we adopt the CLIP ViT-L/14 visual encoder.
At training stage, we initially pre-train the model with the classical cross-entropy loss for sentence generation. Subsequently, we optimize our model using \ours in both reference-free and reference-based versions.
During cross-entropy pre-training, we train our network with the Adam optimizer~\citep{kingma2015adam}, a batch size of 1,024, and for up to 20,000 steps. During this phase, we linearly warmup for 1,000 steps, then keep a constant learning rate of $2.5\cdot10^{-4}$ until 10,000 steps, then sub-linearly decrease until 15,000 steps to $1\cdot10^{-5}$ and keep the value constant until the end of the training.

For the second stage, we further optimize our model using Adam as optimizer with $1\cdot 10^{-6}$ as learning rate, for one epoch using a batch size of 32. During caption generation, we employ a beam size equal to 5. We train our model on the COCO dataset~\citep{lin2014microsoft} using the splits defined by~\cite{karpathy2015deep}.

\begin{table*}[t]
\small
\centering
\caption{Human judgment correlation scores on Flickr8k-Expert and Flickr8k-CF~\citep{hodosh2013framing} and on Composite dataset~\citep{aditya2015images}. The overall best scores are in bold.}
\label{tab:flickr}
\vspace{-0.1cm}
\setlength{\tabcolsep}{.3em}
\resizebox{0.9\linewidth}{!}{
\begin{tabular}{lc cc c cc c cc}
\toprule
& & \multicolumn{2}{c}{\textbf{Flickr8k-Expert}} & & \multicolumn{2}{c}{\textbf{Flickr8k-CF}}  & & \multicolumn{2}{c}{\textbf{Composite}} \\
\cmidrule{3-4} \cmidrule{6-7} \cmidrule{9-10}
& & Kendall $\tau_b$ & Kendall $\tau_c$  & & Kendall $\tau_b$ & Kendall $\tau_c$ & & Kendall $\tau_b$ & Kendall $\tau_c$ \\
\midrule
BLEU-1~\citep{papineni2002bleu} & & 32.2 & 32.3 & & 17.9 & 9.3 & & 29.0 & 31.3 \\
BLEU-4~\citep{papineni2002bleu} & & 30.6 & 30.8 & & 16.9 & 8.7  & & 28.3 & 30.6 \\
ROUGE~\citep{lin2004rouge} & & 32.1 & 32.3 & & 19.9 & 10.3 & & 30.0 & 32.4 \\
METEOR~\citep{banerjee2005meteor} & & 41.5 & 41.8 & & 22.2 & 11.5 & & 36.0 & 38.9 \\
CIDEr~\citep{vedantam2015cider} & & 43.6 & 43.9 & & 24.6 & 12.7 & & 34.9 & 37.7 \\
SPICE~\citep{spice2016} & & 51.7 & 44.9 & & 24.4 & 12.0 & & 38.8 & 40.3 \\
\midrule
BERT-S~\citep{zhang2019bertscore} & & - & 39.2 & & 22.8 & -  & & - & 30.1 \\
LEIC~\citep{cui2018learning} & & 46.6 & - & & 29.5 & - & & - & -\\
BERT-S++~\citep{yi2020improving} & & - & 46.7 & & - & - & & - & 44.9 \\
UMIC~\citep{lee2021umic} & & - & 46.8 & & - & - & & - & -\\
TIGEr~\citep{jiang2019tiger} & & - & 49.3 & & - & - & & - & 45.4 \\
ViLBERTScore~\citep{lee2020vilbertscore} & & - & 50.1 & & - & - & & - & 52.4 \\
MID~\citep{kim2022mutual} & & - & 54.9  & & 37.3 & - & & - & -\\
\midrule
CLIP-S~\citep{hessel2021clipscore} & & 51.1 & 51.2 & & 34.4 & 17.7 & & 49.8 & 53.8 \\

\rowcolor{LightCyan}
& & \underline{54.1} & \underline{54.5} & & \underline{37.0} & \underline{19.1} & & \underline{53.9} & \underline{58.3}  \\
\rowcolor{LightCyan}
\multirow{-2}{*}{\textbf{\ours}} & & (\textcolor{blue}{+3.0}) & (\textcolor{blue}{+3.3}) & & (\textcolor{blue}{+2.6}) & (\textcolor{blue}{+1.4}) & & (\textcolor{blue}{+4.1}) & (\textcolor{blue}{+4.5})\\
\midrule
RefCLIP-S~\citep{hessel2021clipscore} & & 52.6 & 53.0 & & 36.4 & 18.8 & & 51.2 & 55.4 \\
\rowcolor{LightCyan}
& & \underline{\textbf{55.3}} & \underline{\textbf{55.7}} & & \underline{\textbf{37.9}} & \underline{\textbf{19.6}}   & & \underline{\textbf{54.7}} & \underline{\textbf{59.1}}  \\
\rowcolor{LightCyan}
\multirow{-2}{*}{\textbf{\oursref}} & & (\textcolor{blue}{+3.1}) & (\textcolor{blue}{+2.7}) & & (\textcolor{blue}{+1.5}) & (\textcolor{blue}{+0.8}) & & (\textcolor{blue}{+3.5}) & (\textcolor{blue}{+3.7}) \\
\bottomrule
\end{tabular}
}
\vspace{-0.2cm}
\end{table*}

\subsection{Evaluating Human Correlation}
To evaluate the correlation with the human judgment of the proposed metric, we conduct experiments on the Flickr8k-Expert, Flickr8k-CF~\citep{hodosh2013framing}, and Composite~\citep{aditya2015images} for the image setting. Additionally, we employ the VATEX-EVAL dataset~\citep{shi2022emscore} to evaluate video-caption pairs.

\tit{Image Captioning Evaluation} The Flickr8k-Expert and Flickr8k-CF consist of image-caption pairs with the corresponding human ratings. In detail, Flickr8k-Expert comprises 17k expert annotations for visual-textual pairs, with a total of 5,664 distinct images. Each pair receives a score ranging from 1 to 4, where 1 indicates a lack of correlation between the caption and the image, and 4 indicates an accurate depiction of the image without errors. On the other hand, Flickr8k-CF is composed of 145k binary quality judgments, collected from CrowdFlower, covering 48k image-caption pairs that contain 1k unique images. Each pair is annotated with at least three binary scores, where ``yes'' denotes that the caption correlates with the image. We compute the mean proportion of ``yes'' annotations as the score for each pair to measure the alignment with human judgment.

In Table~\ref{tab:flickr}, we report the results comparing the proposed \ours metric with respect to both standard captioning evaluation metrics, such as BLEU~\citep{papineni2002bleu}, METEOR~\citep{banerjee2005meteor}, CIDEr~\citep{vedantam2015cider}, and SPICE~\citep{spice2016})  and more recent solutions, like BERT-S~\citep{zhang2019bertscore}, BERT-S++~\citep{yi2020improving}, TIGEr~\citep{jiang2019tiger}, UMIC~\citep{lee2021umic}, VilBERTScore~\citep{lee2020vilbertscore}, MID~\citep{kim2022mutual}, and CLIP-S~\citep{hessel2021clipscore}.
Only CLIP-S and PAC-S are reported in both reference-free and reference-based versions, while all other metrics require reference captions, except UMIC which is a reference-free evaluation score.

Following previous works~\citep{hessel2021clipscore}, we compute Kendall correlation scores ($\tau_b$ and $\tau_c$). Results reveal that \ours outperforms all other metrics in both the reference-free and reference-based setting. 
Specifically, when comparing our score in a reference-free setting, notable improvements of +3.3 and +2.6 points are observed for $\tau_c$ on Flickr8k-Expert and $\tau_b$ on Flickr8k-CF, respectively. Similar improvements are evident in the reference-based setting. When comparing \ours with standard reference-based metrics such as CIDEr or SPICE, the performance gap widens considerably, reaching +11.7/11.8 points with respect to CIDEr on the Flickr8k-Expert dataset. 

\begin{figure}[t]
    \centering
    \includegraphics[width=0.98\linewidth]{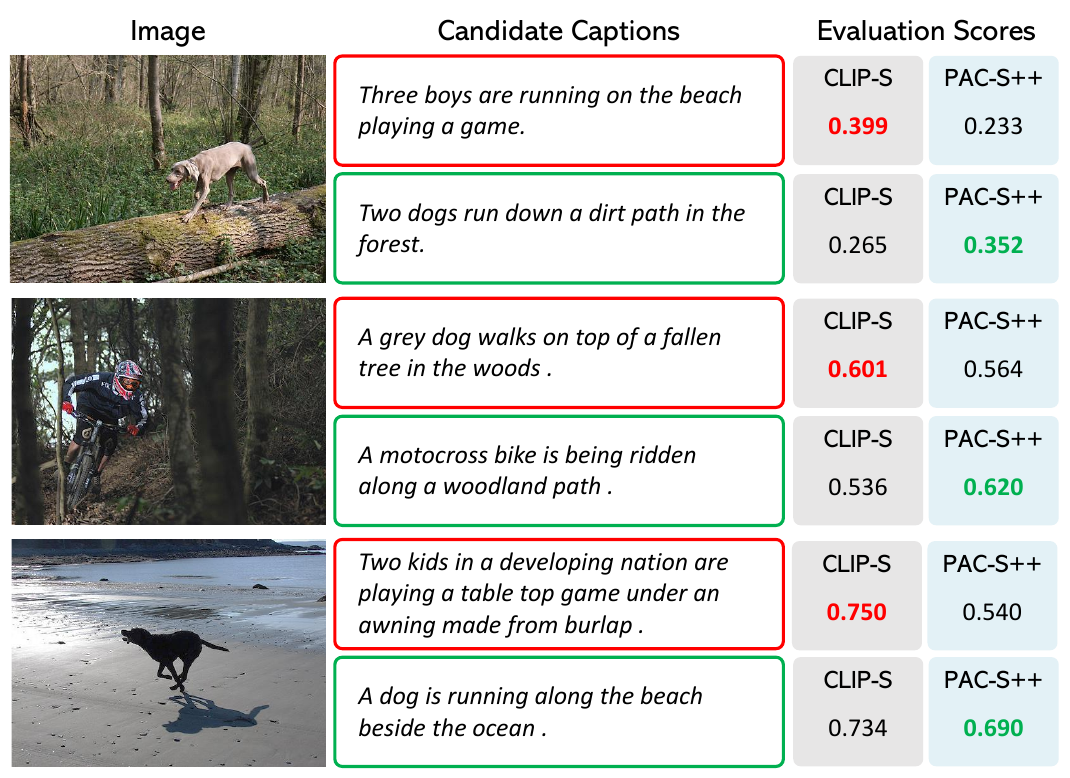}
    \caption{Evaluation scores generated by \ours in comparison with CLIP-S on the Flickr8k-Expert dataset. }
    \vspace{-0.2cm}
    \label{fig:flick_qual}
\end{figure}

Comparable, and even higher, improvements can be noticed in the Composite dataset. This dataset comprises 12,000 human judgments for image-caption pairs, incorporating images taken from COCO~\citep{lin2014microsoft} (2,007 images), Flickr8k~\citep{hodosh2013framing} (997 images), and Flickr30k~\citep{young2014image} (991 images). In this setting, human evaluators were asked to assess each image-caption pair and assign a score within the range of 1 to 5 to estimate the alignment of the caption with the associated image. The results, shown in Table~\ref{tab:flickr}, demonstrate the effectiveness of our metric also in this case, resulting in improvements of +4.5 and +3.7 points in terms of Kendall $\tau_c$ in both reference-free and reference-based settings. 

\begin{table}[t]
\caption{Human judgment correlation scores on VATEX-EVAL dataset~\citep{shi2022emscore} for video captioning evaluation.} 
\label{tab:vatex}
\footnotesize
\vspace{-0.1cm}
\centering
\setlength{\tabcolsep}{.15em}
\begin{tabular}{lc cc c cc c cc}
\toprule
& & \multicolumn{2}{c}{\textbf{No Ref}} & & \multicolumn{2}{c}{\textbf{1 Ref}} & & \multicolumn{2}{c}{\textbf{9 Refs}} \\
\cmidrule{3-4} \cmidrule{6-7} \cmidrule{9-10}
& & $\tau_b$ & $\rho$ & & $\tau_b$ & $\rho$ & & $\tau_b$ & $\rho$ \\
\midrule
BLEU-1 & & - & - & & 12.2 & 15.9 & & 28.9 & 37.0 \\
BLEU-4 & & - & - & & 12.6 & 16.4 & & 22.4 & 29.5 \\
ROUGE & & - & - & & 12.5 & 16.3 & & 23.8 & 30.9 \\
METEOR & & - & - & & 16.4 & 21.5 & & 27.6 & 35.7 \\
CIDEr & & - & - & &  17.3 & 22.6 & & 27.8 & 36.1 \\
\midrule
BERT-S & & - & - & & 18.2 & 23.7 & & 29.3 & 37.8 \\
BERT-S++ & & - & - & & 15.2 & 19.8 & & 24.4 & 31.7 \\
\midrule
EMScore & & 23.2 & 30.3 & & 28.6 & 37.1 & & 36.8 & 47.2 \\
\rowcolor{LightCyan}
\textbf{\ours~/} & & \textbf{28.1} & \textbf{36.4} & & \textbf{32.2} & \textbf{41.5} & & \textbf{39.8} & \textbf{50.8} \\
\rowcolor{LightCyan}
\textbf{\oursref} & & (\textcolor{blue}{+4.9}) & (\textcolor{blue}{+6.1}) & & (\textcolor{blue}{+3.6}) & (\textcolor{blue}{+4.4}) & & (\textcolor{blue}{+3.0}) & (\textcolor{blue}{+3.6})  \\
\bottomrule
\end{tabular}
\vspace{-0.2cm}
\end{table}

Additionally, we present some qualitative results, which are presented in Fig.~\ref{fig:flick_qual}. These results demonstrate that our metric, \ours, exhibits a superior correlation with human judgment compared to the widely used CLIP-S.

\tit{Video Captioning Evaluation} To further validate the robustness of our metric, we compute the correlation with humans in the context of video-caption pairs, employing the VATEX-EVAL dataset. This dataset includes 3k videos from the VATEX~\citep{wang2019vatex} validation set, each of them associated with six captions of mixed quality. Each video-caption pair has been evaluated by three human annotators with a score from 1 (to denote inconsistency between the video and the caption) to 5 (to denote consistency). Overall, the dataset contains 54k human ratings for 18k video-caption pairs. 
Following recently introduced video score~\citep{shi2022emscore}, we compute Kendall $\tau_b$ and Spearman $\rho$ rank correlation coefficients. This evaluation considers varying numbers of reference sentences when measuring correlation, including scenarios with zero, one, or nine references. For instances where no reference is available, our method exhibits noteworthy advancements, achieving increases of +4.9 and +6.1 points in terms of $\tau_b$ and $\rho$ coefficients, respectively, compared to EMScore. These improvements persist across settings with multiple captions, as illustrated in Table~\ref{tab:vatex}.

\begin{table}[t]
\caption{Accuracy results on the Pascal-50S dataset~\citep{vedantam2015cider} obtained by averaging the scores over five random draws of reference captions (except for reference-free metrics). The $\dagger$ marker indicates scores reported in previous works, which may differ in terms of selected reference captions. We refer to the text for the definition of HC, HI, HM, and MM. The overall best scores are in {bold}.}
\label{tab:pascal}
\vspace{-0.1cm}
\footnotesize
\centering
\setlength{\tabcolsep}{.35em}
\begin{tabular}{lc cccc c c}
\toprule
 & & HC & HI & HM & MM & & Mean \\
\midrule
length & & 51.7 & 52.3 & 63.6 & 49.6 & & 54.3 \\
BLEU-1 & & 64.6 & 95.2 & 91.2 & 60.7 & & 77.9 \\
BLEU-4 & & 60.3 & 93.1 & 85.7 & 57.0 & & 74.0 \\
ROUGE & & 63.9 & 95.0 & 92.3 & 60.9 & & 78.0 \\
METEOR & & 66.0 & 97.7 & 94.0 & 66.6 & & 81.1 \\
CIDEr & & 66.5 & 97.9 & 90.7 & 65.2 & & 80.1 \\
\midrule
BERT-S$^\dagger$ & & 65.4 & 96.2 & 93.3 & 61.4 & & 79.1 \\
BERT-S++$^\dagger$ & & 65.4 & 98.1 & 96.4 & 60.3 & & 80.1 \\
TIGEr$^\dagger$ & & 56.0 & 99.8 & 92.8 & 74.2 & & 80.7 \\
ViLBERTScore$^\dagger$ & & 49.9 & 99.6 & 93.1 & 75.8 & & 79.6\\
FAIEr$^\dagger$ & & 59.7 & {\textbf{99.9}} & 92.7 & 73.4 & & 81.4 \\
MID$^\dagger$ & & 67.0 & 99.7 & \textbf{97.4} & \textbf{76.8} & & \textbf{85.2} \\
\midrule
CLIP-S & & 55.9 & 99.3 & 96.5 & 72.0 & & 80.9 \\
\rowcolor{LightCyan}
 & & \underline{59.5} & \underline{99.6} & \underline{96.5} & \underline{73.6} & & \underline{82.3} \\
\rowcolor{LightCyan}
\multirow{-2}{*}{\textbf{\ours}} & & (\textcolor{blue}{+3.6}) & (\textcolor{blue}{+0.3}) & (\textcolor{blue}{+0.0}) & (\textcolor{blue}{+1.6}) & & (\textcolor{blue}{+1.4}) \\
\midrule
RefCLIP-S & & 64.9 & 99.5 & 95.5 & \underline{73.3} & & 83.3 \\
\rowcolor{LightCyan}
& & \underline{\textbf{67.2}} & \underline{99.6} & \underline{96.2} & 74.2 & & \underline{84.5} \\
\rowcolor{LightCyan}
\multirow{-2}{*}{\textbf{\textbf{\oursref}}} & & (\textcolor{blue}{+2.3}) & (\textcolor{blue} {+0.1}) & (\textcolor{blue}{+0.7}) & (\textcolor{blue}{+0.9}) & & (\textcolor{blue}{+1.2}) \\
\bottomrule
\end{tabular}
\vspace{-0.2cm}
\end{table}

\subsection{Caption Pairwise Ranking}
Differently from the datasets presented until now, which include human preferences, the PASCAL-50S dataset~\citep{vedantam2015cider} presents pairwise preference judgments between two captions. This dataset comprises 4k sentence pairs, each associated with an image from the UIUC Pascal sentence dataset~\citep{rashtchian2010collecting}. 
For each pair, 48 human judgments are provided, with each assessment indicating the preferable description for the given image. The sentence pairs are categorized into four groups: (i) both human-written and correct captions (HC), (ii) both human-written captions where one is correct and the other is wrong (HI), (iii) both correct captions but one written by humans and the other machine-generated (HM), (iv) both machine-generated and correct captions (MM).
In this case, where a preference is indicated, we opt for accuracy computation instead of relying on correlation scores. For each caption pair, we compute accuracy considering the caption preferred by the majority of human ratings as correct (with ties resolved randomly). We then assess how often the evaluation metric assigns a higher score to the selected caption. In each evaluation, we conduct random sampling of five reference captions from the pool of 48 provided by the dataset. The results are averaged over five distinct draws. 

From the results presented in Table~\ref{tab:pascal}, we notice that \ours achieves better results than CLIP-S across nearly all categories. These improvements persist also in the reference-based setting, reflecting an average accuracy gain of +1.2 points. In addition to surpassing CLIP-S, our metric also outperforms other standard and more recent metrics, with the only exception of the MID score that, in some categories, attains better accuracy scores.  However, it is important to notice that our results are not directly comparable to those reported in previous works, such as FAIEr and MID. This is due to the random selection of ground-truth sentences when computing reference-based metrics.

\begin{table}[t]
\caption{Accuracy results on the FOIL~\citep{shekhar2017foil} and ActivityNet-FOIL~\citep{shi2022emscore} hallucination detection datasets. The overall best scores are in bold.}
\label{tab:foil}
\vspace{-0.1cm}
\footnotesize
\centering
\setlength{\tabcolsep}{.23em}
\begin{tabular}{lc cc cc}
\toprule
 & & \multicolumn{2}{c}{\textbf{FOIL}} & & \textbf{ActivityNet-FOIL} \\
 \cmidrule{3-4} \cmidrule{6-6}
 & & Accuracy & Accuracy & & Accuracy \\
 & & (1 Ref) & (4 Refs) & & (1 Ref) \\
\midrule
BLEU-1 & & 65.7 & 85.4 & & 60.1 \\
BLEU-4 & & 66.2 & 87.0 & & 66.1 \\
ROUGE & & 54.6 & 70.4 & & 56.7 \\
METEOR & & 70.1 & 82.0 & & 72.9 \\
CIDEr & & 85.7 & 94.1 & & 77.9 \\
MID & & 90.5 & 90.5 & & - \\
\midrule
CLIP-S & & 87.2 & 87.2 & & - \\
EMScore & & - & - & & 89.5 \\
\rowcolor{LightCyan}
& & \underline{90.2} & \underline{90.2} & & \underline{91.0} \\
\rowcolor{LightCyan}
\multirow{-2}{*}{\textbf{\ours}} & & (\textcolor{blue}{+3.0}) & (\textcolor{blue}{+3.0}) & & (\textcolor{blue}{+1.5}) \\
\midrule
RefCLIP-S & & 91.0 & 92.6 & & - \\
EMScoreRef & & - & - & & 92.4 \\
\rowcolor{LightCyan}
& & \underline{\textbf{93.5}} & \underline{\textbf{94.1}} & & \underline{\textbf{93.4}} \\
\rowcolor{LightCyan}
\multirow{-2}{*}{\textbf{\oursref}} & & (\textcolor{blue}{+2.5}) & (\textcolor{blue}{+1.5}) & & (\textcolor{blue}{+1.0}) \\
\bottomrule
\end{tabular}
\vspace{-0.2cm}
\end{table}

\subsection{Sensitivity to Object Hallucination}
Correctly identifying object hallucination in image description is fundamental for the captioning task. Object hallucination refers to the inclusion of objects in the caption that do not actually appear in the corresponding image or video. Therefore, we extend our analysis to two datasets designed for detecting hallucination in textual sentences, namely FOIL~\citep{shekhar2017foil} and ActivityNet-FOIL~\citep{shi2022emscore}. Results about these datasets are reported in Table~\ref{tab:foil}.

\tit{Image Captioning} The FOIL dataset comprises image-caption pairs from the COCO dataset, where captions are modified to introduce a single error, referred to as ``foil word''. For a fair comparison, we select the subset of the validation set that does not overlap with the portion of COCO used during training, resulting in 8,000 images, each paired with a foil-correct textual counterpart. As indicated in the table, \ours significantly outperforms CLIP-S in both the reference-free and reference-based settings. Specifically, without references we observe an improvement of +3.0 points compared to CLIP-S. When considering \oursref, we achieve enhancements of +2.5 and +1.5 points with 1 and 4 references, respectively.

\begin{table*}[t]
\small
\centering
\caption{Comparison with other recent evaluation metrics. The overall best scores are highlighted in bold, while the second-best are underlined.}
\label{tab:others}
\vspace{-0.1cm}
\setlength{\tabcolsep}{.3em}
\resizebox{0.9\linewidth}{!}{
\begin{tabular}{lc cc c c c c c c }
\toprule
&  & & \multicolumn{1}{c}{\textbf{Flickr8k-Expert}} & & \multicolumn{1}{c}{\textbf{Flickr8k-CF}}  & & \multicolumn{1}{c}{\textbf{Composite}} & & \multicolumn{1}{c}{\textbf{Pascal-50S}}\\
\cmidrule{4-4} \cmidrule{6-6} \cmidrule{8-8} \cmidrule{10-10}
& \textbf{Backbone} & & Kendall $\tau_c$  & & Kendall $\tau_b$ & & Kendall $\tau_c$ && Accuracy\\
\midrule
\rowcolor{TitleColor}\multicolumn{7}{l}{\textit{Standard}}  & & & \\
BLEU-4~\citep{papineni2002bleu}  & - & & 30.8 & & 16.9 & & 30.6 && 74.0\\
METEOR~\citep{banerjee2005meteor}  & - & & 41.8 & & 22.2 & & 36.0 & & 81.1 \\
CIDEr~\citep{vedantam2015cider}  & - & & 43.9 & & 24.6 & & 37.7 && 80.1 \\
\midrule
\rowcolor{TitleColor}\multicolumn{7}{l}{\textit{Learnable Unsupervised}}  & & & \\
CLIP-S~\citep{hessel2021clipscore} & CLIP ViT-B/32 & & 51.2 & & 34.4 & & 53.8 && 80.9\\
RefCLIP-S~\citep{hessel2021clipscore} & CLIP ViT-B/32 & & 53.0 & & 36.4 & & 55.4 && 83.3\\
CLIP-S~\citep{hessel2021clipscore} & CLIP ViT-L/14 & & 53.0 & & 35.2 & & 55.4 && 81.7\\
RefCLIP-S~\citep{hessel2021clipscore} & CLIP ViT-L/14 & & 55.7 & & 37.5 & & 56.9 && 84.4\\
\rowcolor{LightCyan}
\textbf{\ours} & CLIP ViT-B/32 & & {54.5} & & {37.0} & & {58.3} & & 82.3  \\
\rowcolor{LightCyan}
\textbf{\oursref} & CLIP ViT-B/32  & & {{55.7}} & & {{37.9}}   & & {{59.1}} & & 84.5 \\
\rowcolor{LightCyan}
\textbf{\ours} & CLIP ViT-L/14 & & 57.4 & & 38.5 & & \underline{62.0} & & 82.4  \\
\rowcolor{LightCyan}
\textbf{\oursref} & CLIP ViT-L/14 & & \textbf{57.9} & & \textbf{38.8} & & 61.6 & & 84.7 \\
\midrule
\rowcolor{TitleColor}\multicolumn{7}{l}{\textit{Learnable Supervised}}  & & & \\
Polos~\citep{wada2024polos} & CLIP ViT-B/16 & & 56.4 & & 37.8 & & 57.6 & & \underline{86.5}\\
\midrule
\rowcolor{TitleColor}\multicolumn{7}{l}{\textit{Additional Components}}  & & & \\
BRIDGE~\citep{sarto2024bridge} & CLIP ViT-L/14 & & 55.8 & & 36.3 & & 57.2 & & 82.9\\
HICE-S~\citep{zeng2024hicescore} & Alpha-CLIP ViT-L/14 & & 56.4 & & 37.2 & & 57.9 & & 86.1\\
RefHICE-S~\citep{zeng2024hicescore} & Alpha-CLIP ViT-L/14 & & \underline{57.7} & & 38.2 & & 58.7 & & \textbf{87.3}\\

\midrule
\rowcolor{TitleColor}\multicolumn{7}{l}{\textit{LLM-based}}  & & & \\
CLAIR~\citep{chan2023clair} & GPT-3.5 & & 48.3 & & - & & 61.0 & & 78.7\\
FLEUR~\citep{lee2024fleur} & LLaVA v1.5-13B & & 53.0 & & \underline{38.6} & & \textbf{63.5} & & 83.2\\
\bottomrule
\end{tabular}
}
\vspace{-0.2cm}
\end{table*}

\tit{Video Captioning} The ActivityNet-FOIL dataset contains video-text pairs from the ActivityNet test set~\citep{zhou2019grounded}. Each video comes with two annotated paragraphs, one used to construct a foil-correct pair and the other used as ground-truth for reference-based metrics. To create a foil caption, a noun phrase in the original caption is replaced with a similar but incorrect visual concept. Overall, the dataset is composed of 1,900 foil-correct paragraph pairs. 
In the video setting, we similarly observe improvements comparable to those in image captioning. Specifically, we observe an improvement of +1.5 and 1.0 points compared to EMScore. These results demonstrate the efficacy of our approach in detecting hallucinations not only in an image-based scenario but also in the case of video sequences.

\subsection{\rev{Comparisons with Advanced Metrics}}
All the competitors cited so far include standard metrics, like BLEU or CIDEr, as well as learnable ones, such as CLIP-S. However, more recent metrics have been introduced in the literature that are not directly comparable to our proposed evaluation score due to significant differences in their training strategies or architectural designs.  Nevertheless, the comparison with these recent metrics is worth mentioning, and the results are presented in Table~\ref{tab:others}, where we show \ours with both CLIP ViT-B/32 and CLIP ViT-L/14 backbones.

\tit{Learnable Supervised Metrics}
Our metric, like other learnable ones, does not train a model to predict a specific score. In contrast, Polos~\citep{wada2024polos} is a supervised metric designed to directly compute evaluation scores by leveraging an annotated dataset and incorporating reference captions as input during training. Although Polos employs a different backbone, our unsupervised training strategy with a ViT-L/14 backbone outperforms Polos, as demonstrated by the higher scores across various datasets. This performance gap is even more pronounced in our reference-based version. These results indicate that a stronger backbone and a better-aligned multimodal embedding space are more effective than directly training to predict a score.

\begin{table*}[t]
\caption{\rev{Evaluation scores of standard and LLM-based captioners on COCO test set.}}
\label{tab:system_level}
\vspace{-0.1cm}
\small
\centering
\setlength{\tabcolsep}{.3em}
\resizebox{0.95\linewidth}{!}{
\begin{tabular}{>{\color{black}}lc >{\color{black}}c>{\color{black}}c>{\color{black}}c>{\color{black}}c>{\color{black}}c>{\color{black}}c>{\color{black}}c c >{\color{black}}c>{\color{black}}c}
\toprule
& & \multicolumn{7}{c}{\rev{\textbf{Reference-based}}} & & \multicolumn{2}{c}{\rev{\textbf{Reference-free}}} \\
\cmidrule{3-9} \cmidrule{11-12}
& & B-4 & M & C & S & RefCLIP-S & Polos & \cellcolor{LightCyan}\textbf{RefPAC-S++} & & CLIP-S & \cellcolor{LightCyan}\textbf{PAC-S++} \\
\midrule
\rowcolor{TitleColor}
\multicolumn{12}{l}{\rev{\textit{Standard Captioning Models}}}\\
Show and Tell~\citep{vinyals2015show} & & 31.4 & 25.0 & 97.2 & 18.1 & 0.779 & 0.585 & \cellcolor{LightCyan}0.752 & & 0.715 & \cellcolor{LightCyan}0.654 \\
Show, Attend and Tell~\citep{xu2015show} & & 33.4 & 26.2 & 104.6 & 19.3 & 0.790 & 0.609 & \cellcolor{LightCyan}0.766 & & 0.727 & \cellcolor{LightCyan}0.670 \\
Up-Down~\citep{anderson2018bottom} & & 36.7 & 27.9 & 122.7 & 21.5 & 0.804 & 0.640 & \cellcolor{LightCyan}0.778 & & 0.740 & \cellcolor{LightCyan}0.680 \\
AoANet~\citep{huang2019attention} & & 39.1 & 29.0 & 128.9 & 22.7 & 0.813 & 0.660 & \cellcolor{LightCyan}0.787 & & 0.753 & \cellcolor{LightCyan}0.693 \\
$\mathcal{M}^2$ Transformer~\citep{cornia2020meshed} & & 39.1 & 29.2 & 131.2 & 22.6 & 0.813 & 0.629 & \cellcolor{LightCyan}0.791 & & 0.757 & \cellcolor{LightCyan}0.699 \\
X-Transformer~\citep{pan2020x} & & 39.7 & 29.5 & 132.8 & 23.4 & 0.819 & 0.668 & \cellcolor{LightCyan}0.792 & & 0.762 & \cellcolor{LightCyan}0.701 \\
DLCT~\citep{luo2021dual} & & 39.8 & 29.5 & 133.8 & 23.0 & 0.821 & 0.670 & \cellcolor{LightCyan}0.795 & & 0.764 & \cellcolor{LightCyan}0.705 \\
COS-Net~\citep{li2022comprehending} & & 42.0 & 30.6 & 141.1 & \textbf{26.6} & 0.829 & 0.692 & \cellcolor{LightCyan}0.803 & & 0.773 & \cellcolor{LightCyan}0.712 \\ 
\midrule
\rowcolor{TitleColor}
\multicolumn{12}{l}{\rev{\textit{LLM-based Models}}}\\
BLIP-2~\citep{li2023blip2} & & \textbf{43.8} & \textbf{31.7} & \textbf{146.0} & 25.2 & 0.838 & \textbf{0.716} & \cellcolor{LightCyan}\textbf{0.810} & & 0.782 & \cellcolor{LightCyan}0.719 \\
IDEFICS-9B~\citep{laurenccon2023obelisc}  & & 36.8 & 28.3 & 125.1 & 22.0 & \textbf{0.840} & 0.699 & \cellcolor{LightCyan}0.803 & & 0.788 & \cellcolor{LightCyan}0.711 \\
LLaVA-1.5-7B~\citep{liu2024improved} & & 8.1 & 28.0 & 69.6 & 21.2 & 0.828 & 0.666 & \cellcolor{LightCyan}0.794 & & 0.785 & \cellcolor{LightCyan}0.707 \\ 
ShareGPT4V~\citep{chen2024sharegpt4v} & & 3.6 & 17.5 & 0.0 & 15.0 & 0.763 & 0.650 & \cellcolor{LightCyan}0.745 & & \textbf{0.827} & \cellcolor{LightCyan}\textbf{0.726} \\
\midrule 
\textit{Humans} & & - & \textit{24.1} & \textit{87.6} & \textit{21.0} & {\textit{0.822}} & \textit{0.654} & \cellcolor{LightCyan}\textit{0.792} & & \textit{0.782} & \cellcolor{LightCyan}\textit{0.710} \\
\bottomrule
\end{tabular}
}
\vspace{-0.4cm}
\end{table*} 

\begin{figure*}[t]
    \centering
    \includegraphics[width=0.99\linewidth]{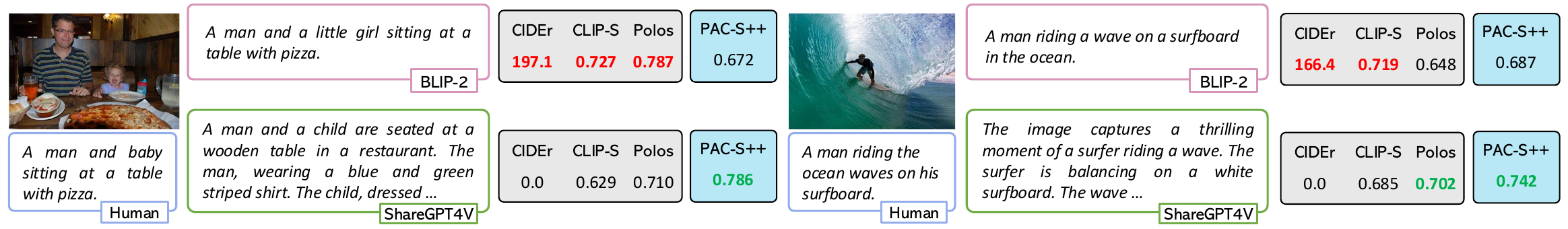}
    \caption{\rev{Qualitative results showing the differences between captions generated by a captioning model trained on COCO captions (\ie, BLIP-2) and LLM-style captions. Quantitative evaluation shows that CIDEr and CLIP-S scores are sensitive to caption length and stylistic variations, while PAC-S++ demonstrates robustness.}}
    \label{fig:qualitatives_metric}
    \vspace{-0.2cm}
\end{figure*}

\tit{Architecturally Enhanced Metrics} 
Another group of methods includes additional components trained for fine-grained evaluation. For instance, the BRIDGE metric~\citep{sarto2024bridge} introduces a mapping module to generate detailed pseudo-captions, aiming for a richer representation. Despite this, our metric, using the same ViT-L/14 architecture, still shows superior performance. Similarly, the HICE-S metric~\citep{zeng2024hicescore} utilizes an interpretable hierarchical scoring mechanism by employing the SAM model~\citep{kirillov2023segment} for mask extraction, which transforms the original image into multiple semantic regions, each with its corresponding masks. 
A specialized CLIP backbone, known as Alpha-CLIP~\citep{sun2024alpha}, 
is then used to process these masks.
Our metric, in both reference-free and reference-based versions, outperforms HICE-S on most datasets, although RefHICE-S achieves slightly better results on the Pascal50-S dataset. This strong performance compared to other methods can be attributed to the innovative hierarchical evaluation design of HICE-S, which aligns more closely with human judgment criteria.

\tit{LLM-based Metrics}
Moreover, some recent metrics take advantage of the extensive pre-training capabilities of Large Language Models (LLMs) to evaluate image-caption pairs. For example, CLAIR~\citep{chan2023clair} exploits the capabilities of the GPT-3.5 model in a training-free setting to evaluate these pairs, while FLEUR~\citep{lee2024fleur} employs the multimodal LLM LLaVA v1.5 which employs a ViT-L/14@336px visual backbone and Vicuna-13B as the language model. Both metrics show strong performance but are generally outperformed by \oursref. The exception is the Composite dataset, where FLEUR achieves the highest score.

Overall, despite differences in training methods, architectural components, and the scale of pre-training, our proposed metrics, \ours and \oursref, consistently deliver the best results. This underscores the robustness and effectiveness of our approach, demonstrating a strong trade-off between efficiency and simplicity.

\subsection{\rev{System-Level Correlation}}
\rev{In Table~\ref{tab:system_level}, we assess whether well established captioning metrics and our proposed one effectively evaluate captions generated by popular existing captioning model and modern LLM-based architectures. Among the standard captioning models, we include Show and Tell~\citep{vinyals2015show} and Show, Attend and Tell~\citep{xu2015show}, Up-Down~\citep{anderson2018bottom}, AoANet~\citep{huang2019attention}, $\mathcal{M}^2$ Transformer~\citep{cornia2020meshed}, X-Transformer~\citep{pan2020x}, DLCT~\citep{luo2021dual}, and the more recently proposed COS-Net model~\citep{li2022comprehending} that incorporates CLIP features.
In addition to reporting evaluations on traditional captioning models, we also include recent LLM-based captioning models, like BLIP-2~\citep{li2023blip2}\footnote{\rev{Note that for BLIP-2 we employ the model fine-tuned on COCO image-caption pairs.}}, IDEFICS-9B~\citep{laurenccon2023obelisc}, LLaVA-1.5-7B~\citep{liu2024improved} and ShareGPTV4~\citep{chen2024sharegpt4v}. 
We evaluate \ours scores against both standard metrics, such as BLEU-4, METEOR, CIDEr, and SPICE, and more recent ones like CLIP-S and Polos. To assess performance, captioning models are compared to a human baseline, where, for each sample, one human-annotated sentence (randomly selected from the five available in the COCO dataset) is used as the candidate caption.} 

\rev{As shown, \ours effectively evaluates human-written captions, assigning scores comparable to those of recent state-of-the-art models like COS-Net and certain LLM-based approaches. In contrast, standard metrics such as METEOR and CIDEr often rank human captions below those generated by weaker models like Show and Tell or Up-Down. This limitation is particularly evident when evaluating captions from LLM-based models -- especially longer ones, such as those produced by ShareGPT4V -- where traditional metrics like CIDEr tend to struggle, as shown in the qualitative examples reported in Fig.~\ref{fig:qualitatives_metric}. In these scenarios, reference-free metrics such as CLIP-S and \ours show a stronger ability to handle longer, LLM-generated captions and provide more reliable evaluations.}

\subsection{Ablation Studies}

\tinytit{Effect of the Scaling Factor $w$}
The scaling factor, denoted by $w$ in Eq.~\ref{eq:clip_score}, is used to adjust the scale of the final metric. This adjustment is made to enhance the numerical readability without impacting the ranking of the results. Notably, CLIP-S proposes this analysis and sets $w=2.5$. In our case, due to the different score distributions, we use different $w$ values when using different backbones, as shown in Fig.~\ref{fig:w_ditribution}. Specifically, all experiments featuring the ViT-B/32 backbone employ $w$ set at $2.5$, while for ViT-L/14, the value of $w$ is set to $3$.

\begin{table*}[t]
\centering
\caption{\rev{Ablation study results of \ours / Ref\ours, using different hyperparameters and synthetic data generators. For each experiment, we report in the last column the averaged improvement compared to the previous version of our metric (\ie~PAC-S~\citep{sarto2023positive}), reported in the first row.}}
\vspace{-0.1cm}
\label{tab:ablations}
\setlength{\tabcolsep}{.2em}
\resizebox{\linewidth}{!}{
\begin{tabular}{>{\color{black}}c>{\color{black}}c>{\color{black}}c c >{\color{black}}c>{\color{black}}c c >{\color{black}}c  c >{\color{black}}c c >{\color{black}}c c >{\color{black}}cc  >{\color{black}}cc >{\color{black}}c c>{\color{black}}c >{\color{black}}c >{\color{black}}c }
\toprule
& & & & \multicolumn{2}{c}{\rev{\textbf{Synthetic Data}}} & & \textbf{Flickr8k-Exp} & & \textbf{Flickr8k-CF} & & \textbf{Composite} & & \textbf{VATEX-EVAL} & & \textbf{Pascal-50S} & & \textbf{FOIL} & & \textbf{ActivityNet-FOIL} &&  \\
\cmidrule{5-6} \cmidrule{8-8} \cmidrule{10-10} \cmidrule{12-12} \cmidrule{14-14} \cmidrule{16-16} \cmidrule{18-18} \cmidrule{20-20} 
\textbf{LoRA $r$} & \textbf{$\lambda_v$} & \textbf{$\lambda_t$} & & Visual & Textual & & Kendall $\tau_c$ & & Kendall $\tau_b$  & & Kendall $\tau_c$ & & Kendall $\tau_b$ &  & Accuracy & & Accuracy & & Accuracy && $\bar{\Delta}$\\
\midrule
- & 0.05 & 0.1 & & SDv1.5 & BLIP & & 53.9 / 55.5 & & 36.0 / 37.6 && 55.7 / 57.3 && 25.1 / 31.4 && 82.4 / 84.7 && 89.9 / 93.7 && 90.1 / 93.5 && \\
\midrule
\midrule
\rowcolor{TitleColor}
\multicolumn{22}{l}{\rev{\textit{Effect of Varying LoRA $r$}}} \\
2 & 0.1 & 0.001 & & SDv1.5 & BLIP & & 54.3 / 55.6 & & 36.9 / 38.0 && 58.2 / 59.1 && 27.4 / {32.3} && 82.2 / 84.4 && 90.1 / 93.5 && 91.1 / 93.6 && +0.71 \\
\rowcolor{LightCyan}
4 & 0.1 & 0.001 & & SDv1.5 & BLIP & & {54.5} / {55.7} && {37.0} / 37.9 && 58.3 / 59.1 &  & {28.1} / 32.2 && {82.3} / {84.5} & & {90.2} / 93.5 && {91.0} / 93.4 && \textbf{+0.78} \\
8 & 0.1 & 0.001 & & SDv1.5 & BLIP && {54.5} / 55.6 & & 36.9 / {38.0} && 58.3 / 59.2 && 27.2 / 32.1 && 82.0 / {84.5} && 90.0 / {93.6} && 90.7 / 93.5 && +0.66 \\
16 & 0.1 & 0.001 & & SDv1.5 & BLIP && 54.3 / 55.5 & & 37.0 / 37.9 && {58.5} / {59.3} && 27.8 / {32.3} && 81.8 / {84.5} && 90.1 / {93.6} && 90.9 / {93.7} && +0.74 \\
\midrule
\rowcolor{TitleColor}
\multicolumn{22}{l}{\rev{\textit{Effect of Varying $\lambda_v$ and $\lambda_t$}}} \\
4 & 0.001 & 0.001 & & SDv1.5 & BLIP & & 54.6 / 55.8 && {37.1} / {38.0} && 57.4 / 58.2 &  & 27.9 / 32.4 && 81.9 / 84.5 & & 90.1 / 93.6 && 90.4 / 93.3 && +0.60 \\
4 & 0.05 & 0.001 & & SDv1.5 & BLIP & & 54.7 / 55.8 && 36.9 / 37.9 && {58.3} / {59.1} &  & 27.5 / 32.0 && 82.1 / 84.5 & & 90.3 / 93.7 && 91.0 / 93.7 && +0.76 \\
\rowcolor{LightCyan}
4 & 0.1 & 0.001 & & SDv1.5 & BLIP & & 54.5 / 55.7 && 37.0 / 37.9 && {58.3} / {59.1} &  & 28.1 / 32.2 && {82.3} / {84.5} & & 90.2 / 93.5 && 91.0 / 93.4 && \textbf{+0.78}\\
4 & 0.5 & 0.001 & & SDv1.5 & BLIP & & 54.4 / 55.8 && 36.8 / 37.8 && 57.9 / 58.9 &  & 27.6 / 31.9 && 81.8 / 84.1 & & 89.8 / 93.6 && 90.4 / 93.4 && +0.53 \\
4 & 0.1 & 0.05 & & SDv1.5 & BLIP & & {54.7} / {55.8} && 37.0 / 38.0 && 58.0 / 58.9 &  & 27.7 / 32.2 && 81.9 / 84.4 & & 90.4 / 93.6 && 90.9 / 93.5 && +0.73 \\
4 & 0.1 & 0.1 & & SDv1.5 & BLIP & & 54.6 / 55.8 && 37.0 / 38.0 && 57.5 / 58.5 &  & 27.7 / 32.0 && 82.3 / 84.3 & & 89.9 / 93.4 && 91.3 / 93.3 && +0.63 \\
\midrule
\rowcolor{TitleColor}
\multicolumn{22}{l}{\rev{\textit{Synthetic Data Contribution}}} \\
4 & 0.1 & 0.001 & & - & - & & 53.9 / 54.9 && 36.7 / 37.7 && 57.5 / 58.4 &  & 26.7 / 31.9 && 81.7 / 83.9 & & 89.8 / 93.2 && 90.1 / 93.2 && +0.20 \\
4 & 0.1 & 0.001 & & - & BLIP & & 54.4 / 55.4 && 36.8 / 37.8 && 57.3 / 58.3 &  & 26.7 / 32.0 && 82.1 / 84.4 & & 89.8 / 93.3 && 90.9 / 93.6 && +0.43 \\
4 & 0.1 & 0.001 & & SDv1.5 & - & & 54.2 / 55.2 && 37.0 / 37.9 && 57.9 / 58.7 &  & 27.0 / 32.1 && 82.0 / 84.4 & & 89.9 / 93.3 && 90.4 / 93.7 && +0.49 \\
\rowcolor{LightCyan}
4 & 0.1 & 0.001 & & SDv1.5 & BLIP & & {54.5} / {55.7} && 37.0 / 37.9 && {58.3} / {59.1} &  & 28.1 / 32.2 && {82.3} / {84.5} & & 90.2 / 93.5 && 91.0 / 93.4 && \textbf{+0.78} \\
\midrule
\rowcolor{TitleColor}
\multicolumn{22}{l}{\rev{\textit{Effect of Varying Synthetic Data Sources}}} \\
\rowcolor{LightCyan}
4 & 0.1 & 0.001 & & SDv1.5 & BLIP & & 54.5 / 55.7 && 37.0 / 37.9 && 58.3 / 59.1 &  & 28.1 / 32.2 && 82.3 / 84.5 & & 90.2 / 93.5 && 91.0 / 93.4 && \textbf{+0.78} \\
4 & 0.1 & 0.001 & & SDv1.5 & IDEFICS-3 & & 54.2 / 55.4 && 36.6 / 37.8 && 58.0 / 58.9 &  & 26.9 / 31.9 && 81.9 / 84.2 & & 89.6 / 93.3 && 90.9 / 93.8 && +0.47 \\
4 & 0.1 & 0.001 & & SDv1.5 & LLaMA-3.2 & & 54.6 / 55.7 && 36.9 / 38.0 && 57.9 / 58.7 &  & 27.1 / 31.9 && 82.1 / 84.3 & & 90.2 / 93.7 && 90.9 / 93.8 && +0.64 \\
4 & 0.1 & 0.001 & & SDv3.5 & BLIP & & 54.7 / 56.0 && 36.8 / 37.9 && 58.1 / 58.9 &  & 27.9 / 31.9 && 82.0 / 84.4 & & 90.2 / 93.4 && 90.9 / 93.6 && +0.71 \\
4 & 0.1 & 0.001 & & FLUX.1 & BLIP & & 54.8 / 56.0 && 37.0 / 38.0 && 58.2 / 59.0 & & 28.0 / 32.0 && 81.8 / 84.4 & & 90.2 / 93.4 && 91.1 / 93.6 && +0.76 \\
\bottomrule
\end{tabular}
}
\vspace{-0.3cm}
\end{table*}

\begin{figure}[t]
\centering
\setlength{\tabcolsep}{.1em}
\begin{tabular}{ccc}
\includegraphics[width=0.325\linewidth]{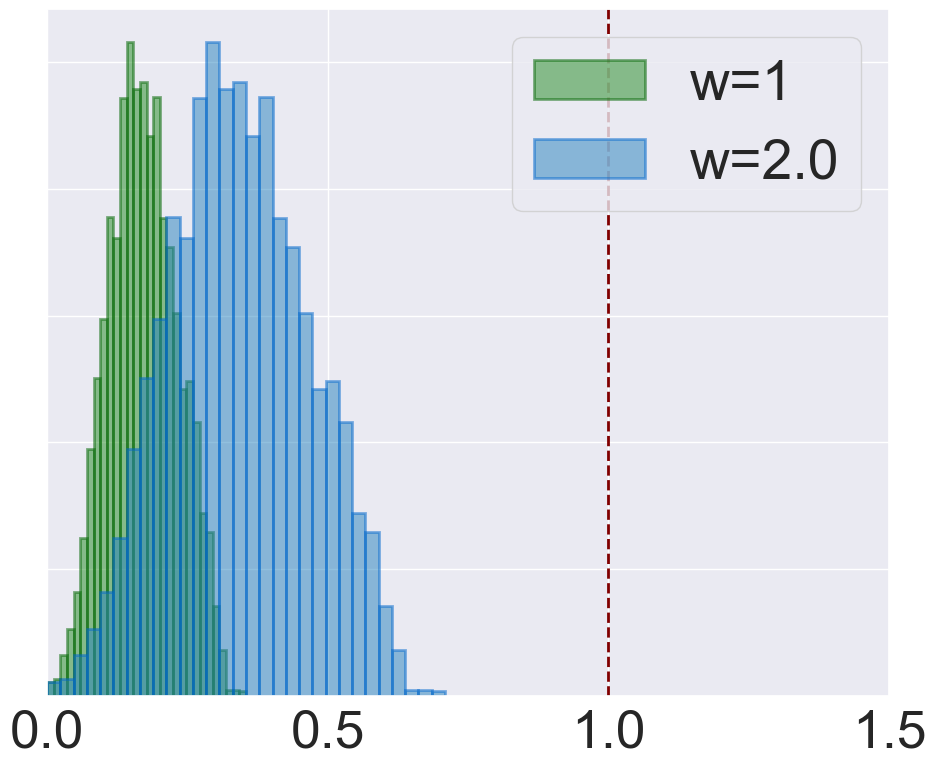} &
\includegraphics[width=0.325\linewidth]{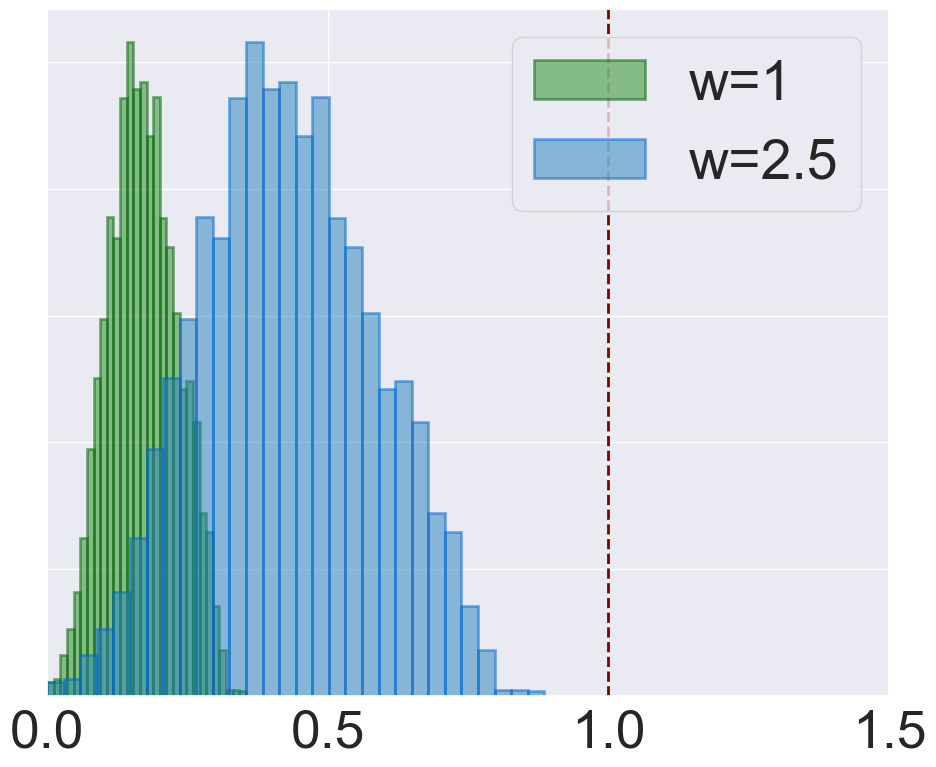} &
\includegraphics[width=0.325\linewidth]{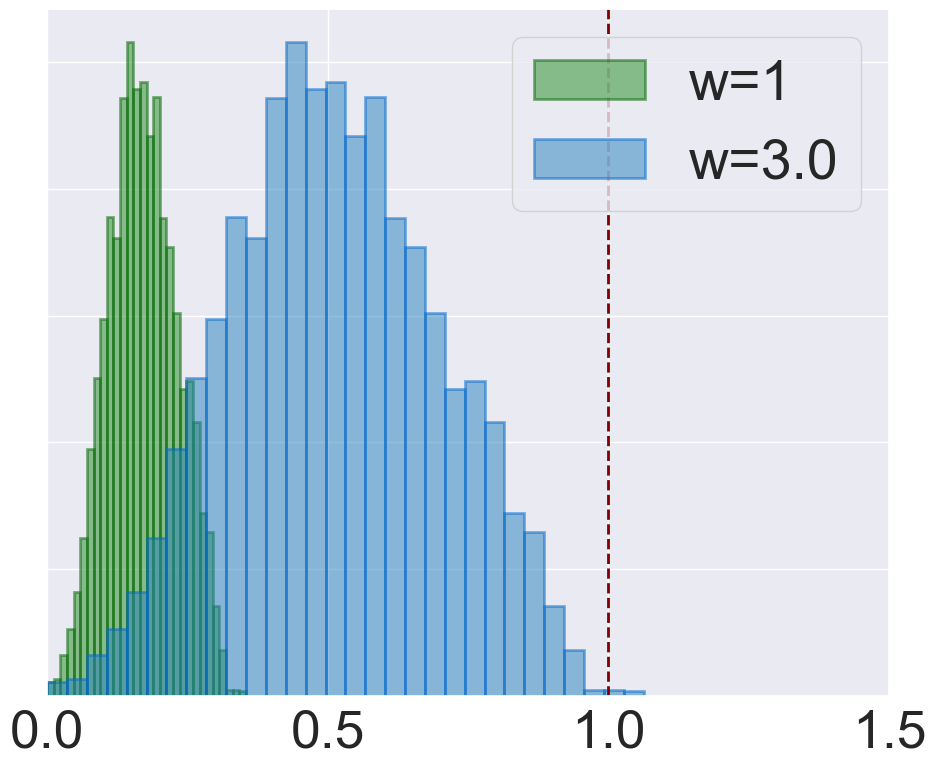} 
\end{tabular}
\hspace{0.5cm}
\begin{tabular}{ccc}
\includegraphics[width=0.325\linewidth]{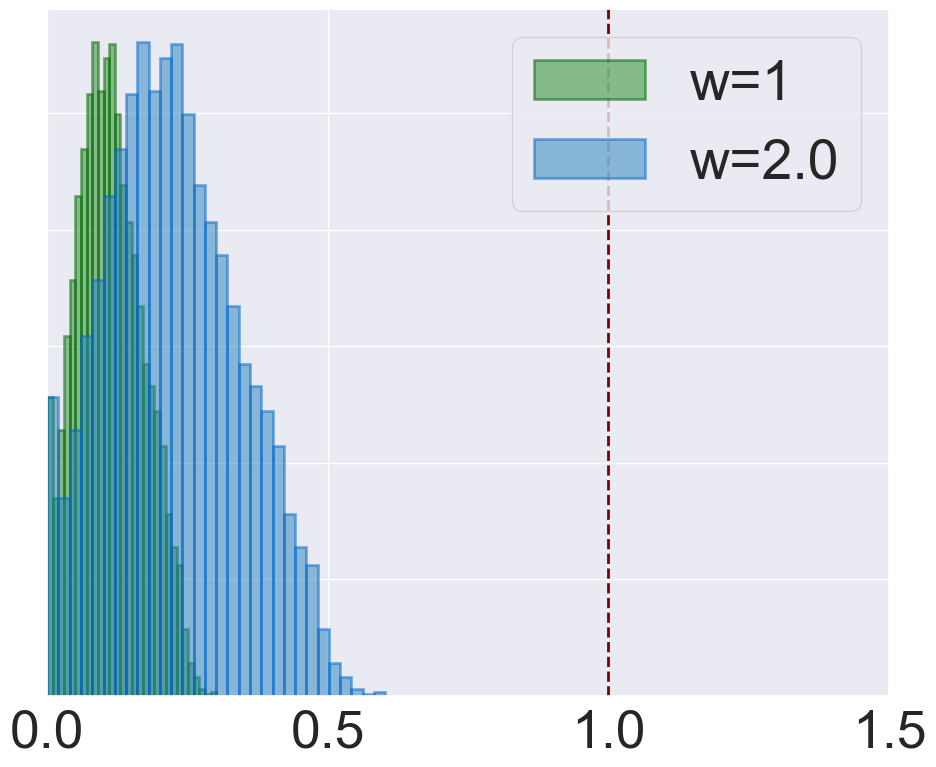} &
\includegraphics[width=0.325\linewidth]{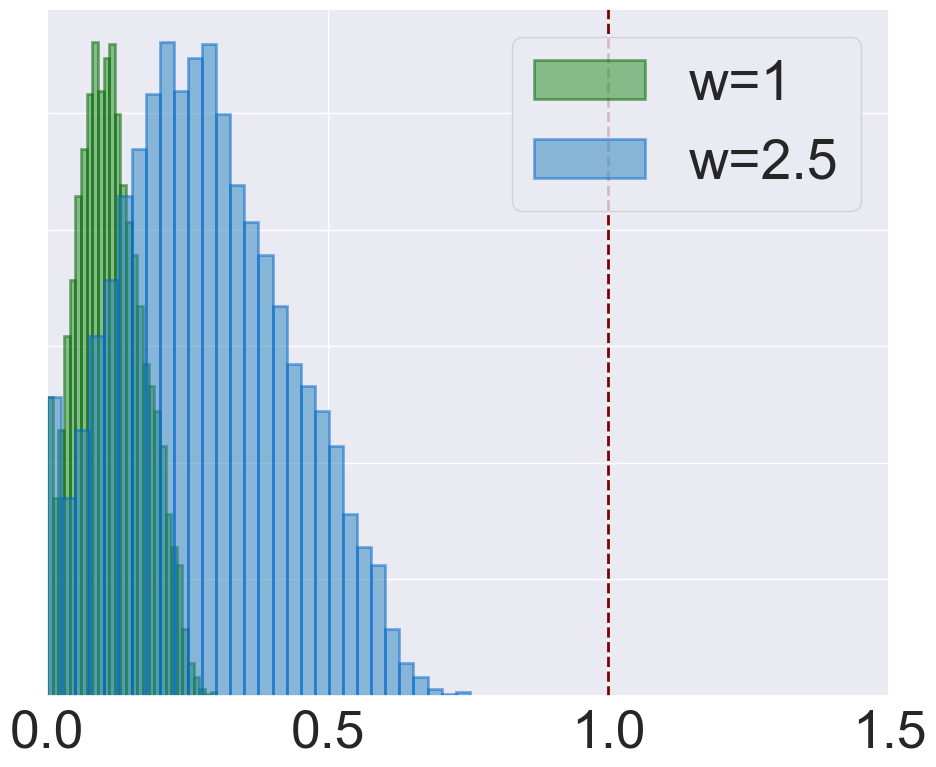} &
\includegraphics[width=0.325\linewidth]{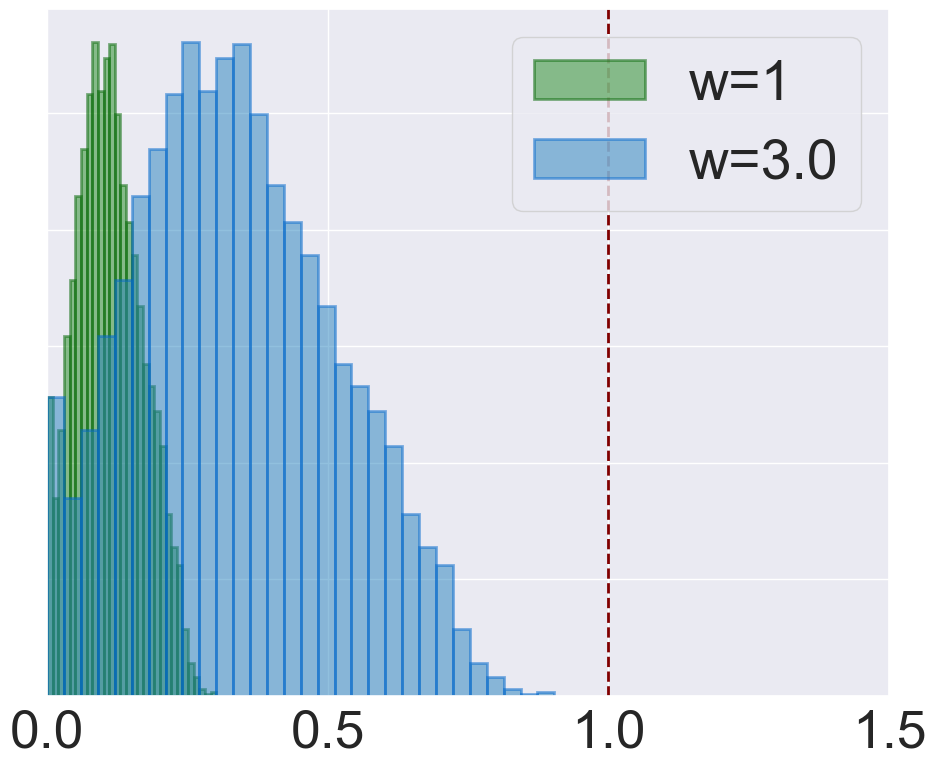} 
\end{tabular}
\caption{Distribution of \ours scores using different scaling factor $w$ on ViT-B/32 (first row) and ViT-L/14 (second row).}
\label{fig:w_ditribution}
\vspace{-.2cm}
\end{figure}

\tit{Low-Rank Analysis}
All the analyses conducted so far employ the \ours version with low-rank adaptation. In Table~\ref{tab:ablations}, we investigate the effect of different ranks (\ie~2, 4, 8, 16) across the selected datasets. Overall, the best performance are achieved with a rank of 4, in both reference-free and reference-based settings. 

Notably, comparing the results of the previous version of our metric (\ie~PAC-S~\citep{sarto2023positive}, first row), in which only the last visual and textual projections of the model are fine-tuned, the version of our metric with LoRA consistently outperforms the original version regardless of the rank, with the only exception of Pascal-50S and FOIL datasets. For example, employing \ours with the CLIP ViT-B/32 backbone fine-tuned with LoRA yields superior results compared to its counterpart without LoRA, achieving 58.3 on the Composite dataset (+2.6) and 28.1 on the VATEX-EVAL dataset (+3.0). This demonstrates the effectiveness of this strategy, even in the context of video settings.

\tit{\rev{Choice of Hyperparameters}}
\rev{Subsequent to determining the optimal rank dimension as $r=4$, we also conduct a comprehensive grid search to determine the optimal values of the $\lambda_v$ and $\lambda_t$ hyperparameters for our model across multiple datasets. The results are summarized in Table~\ref{tab:ablations}. While the results show notable variation in performance depending on the dataset, we observe that the configuration corresponding $\lambda_v=0.1$ and $\lambda_t=0.001$ consistently yields strong improvements across most datasets. Accordingly, we adopt this setting as our final configuration, which is used in the loss function defined in Eq.~\ref{eq:final_loss}.}

\begin{table*}[t]
\centering
\caption{\rev{Human correlation and accuracy results when varying visual and textual backbones. The best results for each backbone are highlighted in bold.}}
\label{tab:backbones}
\vspace{-0.1cm}
\setlength{\tabcolsep}{.35em}
\resizebox{0.92\linewidth}{!}{
\begin{tabular}{lccc cc cc cc cc cc cc}
\toprule
 & & & & \multicolumn{2}{c}{\textbf{Flickr8k-Exp}} & \multicolumn{2}{c}{\textbf{Flickr8k-CF}} & \multicolumn{2}{c}{\textbf{Composite}} & \textbf{Pascal-50S} & & \textbf{FOIL} \\
 \cmidrule(lr){5-6} \cmidrule(lr){7-8} \cmidrule(lr){9-10} \cmidrule(lr){11-11} \cmidrule(lr){13-13}
\textbf{Backbone} & & & & Kendall $\tau_b$ & Kendall $\tau_c$ & Kendall $\tau_b$ & Kendall $\tau_c$ & Kendall $\tau_b$ & Kendall $\tau_c$ & Accuracy & & Accuracy \\
\midrule
 & & CLIP-S && 52.6 & 53.0  &  35.2 & 18.2 & 51.3 & 55.4  & 81.7 && 90.9 \\
 & & PAC-S  && 55.1 & 55.5 & {36.8} & {19.0} & {52.3} & {56.5} & 82.2 && 91.9 \\
\rowcolor{LightCyan}
\cellcolor{white} & \cellcolor{white} & \textbf{\ours} && 57.0 & \underline{57.4}  & \underline{38.5} & 19.9. & 57.3 & \textbf{\underline{62.0}} & \underline{82.4} && \underline{93.6} \\
\cmidrule{3-13}
 & & RefCLIP-S && 54.1 & 54.5 & 36.5 & 18.9 & 51.9 & 56.1 & 84.3 && 94.0 \\
 & & RefPAC-S && 56.7 & 57.1 & 37.7 & 19.5 & 53.1 & 57.2 & 85.0 && \textbf{\underline{95.3}} \\
 \rowcolor{LightCyan}
\cellcolor{white}\multirow{-6}{*}{\textbf{CLIP ViT-L/14}} & \cellcolor{white} & \textbf{\oursref} && \underline{57.5} & \textbf{\underline{57.9}} & \textbf{\underline{38.8}} & \underline{20.1} & \underline{60.0} & \underline{61.6} & \textbf{\underline{85.4}} && 95.1 \\
\midrule
 & & \rev{CLIP-S} && \rev{0.49} & \rev{0.49} & \rev{-1.1} & \rev{-0.6} & \rev{26.1} & \rev{28.2} & \rev{62.2} && \rev{52.5}
\\
 & & \rev{PAC-S} && \rev{58.8} & \rev{\underline{54.0}} & \rev{39.7} & \rev{\textbf{19.9}} & \rev{54.5} & \rev{58.9} & \rev{81.7} && \rev{92.3} \\
\rowcolor{LightCyan}
\cellcolor{white} & \cellcolor{white} & \rev{\textbf{\ours}} && \rev{\underline{61.3}} & \rev{{42.7}} & \rev{\textbf{48.5}} & \rev{19.7} & \rev{\textbf{58.8}} & \rev{\underline{61.5}} & \rev{\underline{81.8}} && \rev{\underline{94.1}}
 \\
\cmidrule{3-13}
 & & \rev{RefCLIP-S} && \rev{33.3} & \rev{33.6} & \rev{18.8} & \rev{9.7} & \rev{30.9} & \rev{33.4} & \rev{73.0} && \rev{69.8} \\
 & & \rev{RefPAC-S} && \rev{59.0} & \rev{\textbf{54.2}} & \rev{40.0} & \rev{\textbf{20.0}} & \rev{55.1} & \rev{59.5} & \rev{83.1} && \rev{93.6} \\
 \rowcolor{LightCyan}
\cellcolor{white}\multirow{-6}{*}{\rev{\textbf{SigLIP ViT-L/14}}} & \cellcolor{white} & \rev{\textbf{\oursref}} && \rev{\textbf{61.4}} & \rev{42.8} & \rev{\textbf{48.5}} & \rev{19.7} & \rev{\textbf{58.8}} & \rev{\textbf{61.6}} & \rev{\textbf{82.7}} && \rev{\textbf{94.4}} \\
\midrule
 & & \rev{CLIP-S} && \rev{4.6} & \rev{4.5} & \rev{1.2} & \rev{0.6} & \rev{19.6} & \rev{21.2} & \rev{59.3} && \rev{61.0} \\
 & & \rev{PAC-S} && \rev{4.5} & \rev{4.1} & \rev{1.1} & \rev{0.5} & \rev{8.6} & \rev{8.6} & \rev{54.9} && \rev{41.2} \\
\rowcolor{LightCyan}
\cellcolor{white} & \cellcolor{white} & \rev{\textbf{\ours}} && \rev{\underline{60.1}} & \rev{\underline{47.4}} & \rev{\textbf{42.6}} & \rev{\textbf{19.3}} & \rev{\underline{59.1}} & \rev{\underline{62.6}} & \rev{\underline{81.0}} && \rev{\underline{93.4}} \\
\cmidrule{3-13}
 & & \rev{RefCLIP-S} && \rev{4.7} & \rev{4.6} & \rev{1.2} & \rev{0.4} & \rev{29.9} & \rev{32.4} & \rev{72.7} && \rev{77.3} \\
 & & \rev{RefPAC-S} && \rev{4.5} & \rev{4.2} & \rev{1.0} & \rev{0.5} & \rev{9.7} & \rev{9.7} & \rev{55.3} && \rev{43.4} \\
 \rowcolor{LightCyan}
\cellcolor{white}\multirow{-6}{*}{\rev{\textbf{SigLIP2 ViT-L/14}}} & \cellcolor{white} & \rev{\textbf{\oursref}} && \rev{\textbf{60.2}} & \rev{\textbf{47.5}} & \rev{\textbf{42.6}} & \rev{\textbf{19.3}} & \rev{\textbf{59.2}} & \rev{\textbf{62.7}} & \rev{\textbf{81.6}} && \rev{\textbf{93.9}} \\
\bottomrule
\end{tabular}
}
\vspace{-0.2cm}
\end{table*}

\tit{\rev{Synthetic Data Generator Analysis}}
\rev{To assess the overall contribution of synthetic data and the impact of different text and image generators, we conduct an ablation study, always reported in Table~\ref{tab:ablations}. We start by analyzing the contribution of each generator. Specifically, we conduct experiments without synthetic data (\ie~only using pairs from COCO during fine-tuning), as well as analyze the effects of removing only the textual or the visual positive augmentations. The averaged improvements compared to the previous version of the metric highlight that both visual and textual synthetic data contribute to enhanced performance compared to the version without any positive augmentation. Furthermore, when both augmentations are applied simultaneously, further performance improvement is observed.}

\rev{As additional analysis, we replace synthetic data generators with more recent models to assess their impact on performance. For text generation, we compare our default setup, which employs BLIP~\citep{li2022blip}, with LLaMA 3.2~\citep{dubey2024llama} and IDEFICS-3~\citep{laurenccon2024building} as caption generators. For image generation, instead, we assess Stable Diffusion v3.5 (SDv3.5)\footnote{\tiny{\rev{https://huggingface.co/stabilityai/stable-diffusion-3.5-large}}} and FLUX\footnote{\tiny{\rev{https://huggingface.co/black-forest-labs/FLUX.1-dev}}} as alternatives to our baseline, Stable Diffusion V1.5 (SDv1.5).
Our analysis indicates that the original configuration remains the most effective overall. In particular, on the text side, substituting newer models does not yield noticeable performance improvements. This can be due to the behavior of modern text generators, which tend to produce longer and more verbose captions. CLIP textual encoder often truncates these, thereby limiting their effective contribution. On the image side, although newer generators may offer improved visual quality, results are comparable with the initial version using SDv1.5 images, overall confirming the effectiveness of our positive-augmented fine-tuning strategy.}

\tit{\rev{Effect of Changing Backbone}}
\rev{We investigate the impact of changing the backbone architecture by replacing the standard CLIP visual and textual encoders with those from SigLIP~\citep{zhai2023sigmoid} and SigLIP2~\citep{tschannen2025siglip2multilingualvisionlanguage}, which differ from CLIP in both training data and training strategies. To ensure a fair comparison, we use the same ViT-L/14 backbone for all three models and we report results in Table~\ref{tab:backbones}. For each, we evaluate both the reference-free and reference-based variants, applying the standard CLIP-S setup (analogous to a zero-shot setting), as well as PAC-S and our PAC-S++ training strategies. Our results show that CLIP-S with the CLIP ViT-L/14 backbone, while underperforming compared to PAC-S and PAC-S++, still achieves competitive results. However, when using SigLIP or SigLIP2, CLIP-S performance significantly degrades, particularly in terms of human correlation, pairwise ranking, and hallucination detection. This suggests that the zero-shot capabilities of SigLIP-like models are less aligned with the image captioning evaluation tasks.
By applying the PAC-S training strategy to SigLIP, we observe substantial improvements, often surpassing even the performance of \ours with the CLIP ViT-L/14 backbone across several datasets. Further gains are achieved using the \ours strategy. For instance, with the SigLIP backbone, Ref\ours achieves $\tau_c$ scores of 61.4 and 58.8 on the Flickr8k-Expert and Composite datasets, respectively, which are significantly higher than the CLIP-based version. These results highlight the importance of both the model architecture and the fine-tuning strategy in achieving robust performance across captioning evaluation benchmarks.
}

\subsection{PAC-Score++ for RL-based Captioning Fine-tuning}
We then evaluate the effectiveness of the proposed \ours metric when employed as reward for fine-tuning a captioning model, using the fine-tuning strategy described in Sec.~\ref{sec:SCST}. In this setting, we compare our metric in both its reference-free and reference-based versions respectively against CLIP-S and RefCLIP-S. For completeness, we also report the results of the model trained with cross-entropy loss only (\ie~without reinforcement learning) and using the standard CIDEr score as reward. \rev{To evaluate generated captions, we employ a combination of traditional metrics, like BLEU, METEOR, CIDEr, and SPICE, and more recent ones such as CLIP-S, Polos and the proposed \ours metric, considering in both cases reference-based and reference-free settings.}
Additionally, we introduce novel metrics to assess the grammatical correctness of the generated captions, which is crucial especially when directly optimizing CLIP-based scores. Specifically, we measure the average number of repeated $n$-grams (Rep-$n$) and the percentage of captions ending with undesirable words like prepositions, conjunctions, or determiners (\%Incorrect).

\tit{In-domain Evaluation} Captioning results on the COCO test set are reported in Table~\ref{tab:coco}. Notably, although CLIP remains an excellent model for aligning bag-of-words with visual input, it disregards syntax and logical connections among words within captions. On the contrary, despite sharing the same architecture, our proposal mitigates this issue, favouring the use of \ours as a reward metric in a captioning model. 
In particular, directly optimizing CLIP-S leads to protracted and repetitive captions, as demonstrated by the lower scores in terms of standard reference-based metrics and grammar measures. In contrast, \ours significantly stabilizes the fine-tuning process, yielding significant enhancements in reference-based metrics (\eg~36.3 and 51.8 CIDEr points using \ours with ViT-B/32 and ViT-L/14 features vs. 1.1 and 1.4 obtained by CLIP-S). Concurrently, it enables the generation of semantically rich and grammatically correct captions that better correlate with human-generated content. This phenomenon is particularly notable in repetitiveness metrics, where the average number of repeated 1-grams in the generated captions decreases from $11.225$ to $4.157$, when using VIT-L/14 as visual backbone. 

\begin{table*}[t]
  \centering
    \caption{Captioning results in terms of reference-based, reference-free, and grammar evaluation metrics on COCO test set, using visual features extracted from different CLIP-based backbones as input to the captioning model.}
  \label{tab:coco}
  \vspace{-0.1cm}
  \setlength{\tabcolsep}{.25em}
  \resizebox{\linewidth}{!}{
  \begin{tabular}{lc c ccccc>{\color{black}}cc c cc c ccccc}
    \toprule
    & & & \multicolumn{7}{c}{\textbf{Reference-based} $\uparrow$} & & \multicolumn{2}{c}{\textbf{Reference-free} $\uparrow$} & & \multicolumn{5}{c}{\textbf{Grammar} $\downarrow$} \\
    \cmidrule{4-10} \cmidrule{12-13} \cmidrule{15-19}
    \textbf{Backbone} & \textbf{Reward} & & B-4 & M & C & S & RefCLIP-S & Polos & RefPAC-S++ & & CLIP-S & PAC-S++ & & Rep-1 & Rep-2 & Rep-3 & Rep-4 & \%Incorrect \\
    \midrule
    & - & & 33.1 & 28.2 & 112.4 & 20.5 & 0.804 & 0.651 & 0.794 & & 0.755  & 0.712 & & 1.468 & 0.091 & 0.017 & 0.005 & 0.3 \\
    & CIDEr & & 40.4 & 29.4 & 129.6 & 21.6 & 0.806 & 0.651 & 0.799 & & 0.751 & 0.714 & & 1.318 & 0.038 & 0.006 & 0.004 & 24.7 \\
    \cmidrule{2-19}
    & CLIP-S & & 12.1 & 23.5 & 1.1 & 20.0 & 0.767 & 0.635 & 0.776 & & \textbf{0.844} & 0.744 & & 12.226 & 4.736 & 1.884 & 0.795 & 99.2 \\
    \rowcolor{LightCyan}
    \cellcolor{white} & \textbf{\ours} & & \textbf{19.4} & \textbf{27.1} & \textbf{36.3} & \textbf{22.4} & \textbf{0.801} & \textbf{0.658} & \textbf{0.795} & & 0.813 & \textbf{0.755} & & \textbf{5.129} & \textbf{1.431} & \textbf{0.544} & \textbf{0.229} & \textbf{0.7} \\
    \cmidrule{2-19}
    & RefCLIP-S & & 26.3 & 27.6 & 92.5 & 21.4 & \textbf{0.829} & \textbf{0.679} & 0.807 & & \textbf{0.799} & 0.735 & & 2.571 & 0.626 & 0.236 & 0.103 & \textbf{0.3} \\
    \rowcolor{LightCyan}
    \cellcolor{white}\multirow{-6}{*}{\textbf{ViT-B/32}} & \textbf{Ref\ours} & & \textbf{30.5} & \textbf{28.5} & \textbf{109.1} & \textbf{22.2} & 0.822 & 0.677 & \textbf{0.811} & & 0.784 & \textbf{0.740} & & \textbf{1.791} & \textbf{0.247} & \textbf{0.069} & \textbf{0.026} & \textbf{0.3} \\
    \midrule
    & - & & 34.8 & 29.9 & 119.4 & 22.5 & 0.802 & 0.078 & 0.708 & & 0.749 & 0.708 & & 1.469 & 0.064 & 0.008 & 0.002 & 0.3 \\
    & CIDEr & & 43.6 & 30.8 & 143.3 & 23.2 & 0.809 & 0.668 & 0.804 & & 0.750  & 0.713 & & 1.432 & 0.047 & 0.005 & 0.002 & 32.3 \\
    \cmidrule{2-19}
    & CLIP-S & & 13.1 & 24.6 & 1.4 & 20.0 & 0.782 & 0.656 & 0.780 & & \textbf{0.840} & 0.736 & & 11.225 & 4.447 & 2.08 & 1.012 & 34.8 \\
    \rowcolor{LightCyan}
    \cellcolor{white} & \textbf{\ours} & & \textbf{20.9} & \textbf{28.0} & \textbf{51.8} & \textbf{23.9} & \textbf{0.806} & \textbf{0.675} & \textbf{0.797} & & 0.812 & \textbf{0.751} & & \textbf{4.157} & \textbf{0.974} & \textbf{0.33} & \textbf{0.129} & \textbf{1.3} \\
    \cmidrule{2-19}
    & RefCLIP-S & & 27.8 & 28.8 & 101.9 & 23.3 & \textbf{0.833} & 0.700 & 0.811 & & \textbf{0.800}  & 0.734 & & 2.161 & 0.386 & 0.13 & 0.046 & 0.7 \\
    \rowcolor{LightCyan}
    \cellcolor{white}\multirow{-6}{*}{\textbf{ViT-L/14}} & \textbf{Ref\ours} & & \textbf{32.5} & \textbf{29.6} & \textbf{118.9} & \textbf{23.5} & 0.826 & \textbf{0.702} & \textbf{0.814} & & 0.782 & \textbf{0.736} & & \textbf{1.468} & \textbf{0.145} & \textbf{0.037} & \textbf{0.011} & \textbf{0.9} \\
    \bottomrule
  \end{tabular}
}
\vspace{-0.6cm}
\end{table*}

\begin{figure*}[t]
    \centering
    \includegraphics[width=0.98\linewidth]{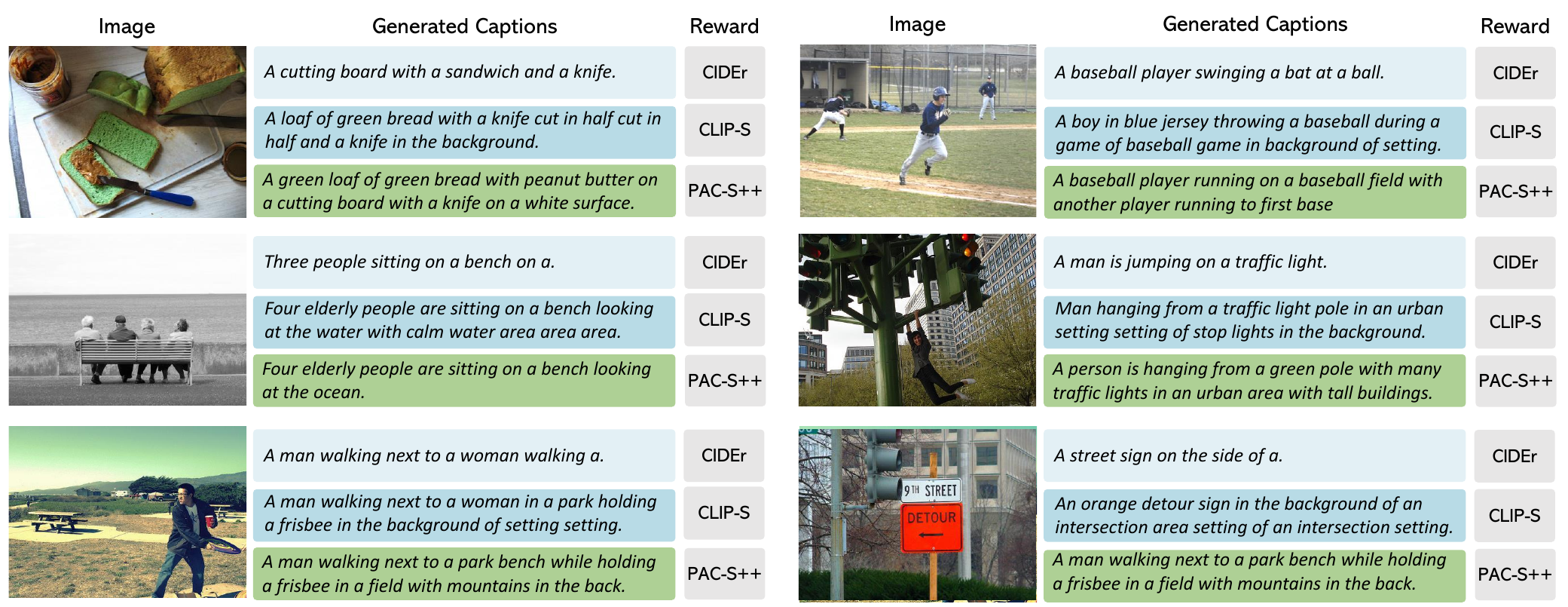}
    \caption{Qualitative image captioning results employing different metrics as reward.}
    \label{fig:reward_qualitatives}
    \vspace{-0.3cm}
\end{figure*} 

Similar considerations apply to the reference-based version, where a reduction in caption generation creativity is observed to align more closely with ground-truth sentences. This approach results in a softer degradation of reference-based metrics, producing values nearly identical to those obtained by the baseline model trained with cross-entropy loss, but achieving higher scores in learnable metrics (\eg~0.708 and 0.713 in terms of \ours respectively with cross-entropy loss only and CIDEr as reward vs. 0.736 achieved when employing Ref\ours as a reward).

To validate the quality of generated captions, qualitative results on sample images from the COCO dataset are reported in Fig.~\ref{fig:reward_qualitatives}, where we compare captions generated by the model fine-tuned using \ours as reward with those generated using CIDEr or CLIP-S. As it can be seen, our proposal can generate more descriptive and detailed captions, while reducing repetitions and grammatical errors. Specifically, while CIDEr generally leads to shorter captions, both CLIP-S and \ours can comprehensively describe the visual content of the images. At the same time, however, using CLIP-S as reward significantly reduces the grammatical correctness of generated captions. This drawback is consistently mitigated when employing \ours as reward, further demonstrating the effectiveness of our solution.

\tit{Out-of-domain Evaluation}
Finally, we evaluate the out-of-domain performance of our model on the nocaps~\citep{agrawal2019nocaps} and VizWiz~\citep{gurari2020captioning} datasets, both of which present distinct image descriptions compared to the COCO dataset used for training. Specifically, the nocaps dataset, which is designed for the novel object captioning task, includes image-caption pairs featuring objects not present in the COCO training set. In contrast, VizWiz consists of images taken by visually impaired individuals, often showcasing challenging perspectives, such as close-up shots or unconventional viewpoints. The results, summarized in Table~\ref{tab:zeroshot}, are evaluated using both reference-free and reference-based metrics.
Also in these challenging settings, our approach demonstrates greater semantic richness while preserving fluidity and grammatical correctness in text generation. This behavior is not observed when CLIP-S is used as reward. Specifically, although the use of CLIP-S results in high scores on learnable metrics, the values of traditional metrics remain notably low. For instance, on the nocaps dataset and using VIT-L/14 as visual backbone, the CIDEr score drops from 49.1 points when using \ours as reward to just 2.1 points with CLIP-S as reward, further highlighting the advantages of our proposed metric for training captioning models. 

\begin{table*}[t]
  \centering
    \caption{Captioning results in terms of reference-based and reference-free metrics on nocaps and VizWiz validation sets.}
  \label{tab:zeroshot}
  \vspace{-0.1cm}
  \setlength{\tabcolsep}{.25em}
  \resizebox{\linewidth}{!}{
  \begin{tabular}{lc c cccc>{\color{black}}cc c cccc>{\color{black}}cc}
    \toprule
    & & & \multicolumn{6}{c}{\textbf{nocaps}} & & \multicolumn{6}{c}{\textbf{VizWiz}} \\
    \cmidrule{4-9} \cmidrule{11-16}
    \textbf{Backbone} & \textbf{Reward} & & C & CLIP-S & PAC-S++ & RefCLIP-S & Polos & RefPAC-S++ & & C & CLIP-S & PAC-S++ & RefCLIPS & Polos & RefPAC-S++ \\
    \midrule
    & - & & 67.6 & 0.686 & 0.694 & 0.699 & 0.432 & 0.733 & & 27.8 & 0.655 & 0.675 & 0.691 & 0.360 & 0.729 \\
    & CIDEr & & 76.2 & 0.695 & 0.703 & 0.709 & 0.418 & 0.741 & & 28.9 & 0.663 & 0.686 & 0.704 & 0.345 & 0.739 \\
    \cmidrule{2-16}
    & CLIP-S & & 1.6 & 0.780 & 0.726 & 0.675 & 0.467 & 0.724 & & 1.1 & \textbf{0.735} & 0.703 & 0.686 & 0.403 & 0.722 \\
    \rowcolor{LightCyan}
    \cellcolor{white}  & \textbf{\ours} & & \textbf{34.6} & \textbf{0.751} & \textbf{0.743} & \textbf{0.713} & \textbf{0.485} & \textbf{0.748} & & \textbf{17.5} & 0.721 & \textbf{0.729} & \textbf{0.717} & \textbf{0.434} & \textbf{0.751} \\
    \cmidrule{2-16}
    & Ref-CLIP-S & & 64.0 & \textbf{0.736} & 0.724 & \textbf{0.734} & \textbf{0.475} & 0.753 & & 25.0 & \textbf{0.703} & 0.708 & \textbf{0.723} & 0.405 & 0.747 \\
    \rowcolor{LightCyan}
    \cellcolor{white}\multirow{-6}{*}{\textbf{ViT-B/32}} & \textbf{\oursref} & & \textbf{73.1} & 0.724 & \textbf{0.729} & 0.728 & 0.474 & \textbf{0.758} & & \textbf{29.4} & 0.694 & \textbf{0.715} & \textbf{0.723} & \textbf{0.412} & \textbf{0.758} \\
    \midrule
    & - & & 75.2 & 0.698 & 0.704 & 0.710 & 0.473 & 0.743 & & 35.0 & 0.655 & 0.679 & 0.701 & \textbf{0.407} & 0.740 \\
    & CIDEr & & 91.3 & 0.698 & 0.711 & 0.718 & 0.438 & 0.755 & & 39.6 & 0.667 & 0.683 & 0.722 & 0.365 & 0.751 \\
    \cmidrule{2-16}
    & CLIP-S & & 2.1 & \textbf{0.791} & 0.741 & 0.705 & 0.508 & 0.746 & & 1.6 & \textbf{0.727} & 0.703 & 0.711 & 0.427 & 0.741 \\
    \rowcolor{LightCyan}
    \cellcolor{white}  & \textbf{\ours} & & \textbf{49.1} & 0.769 & \textbf{0.754} & \textbf{0.735} & \textbf{0.536} & \textbf{0.764} & & \textbf{26.1} & 0.713 & \textbf{0.723} & \textbf{0.726} & \textbf{0.474} & \textbf{0.759} \\
    \cmidrule{2-16}
    & Ref-CLIP-S & & 79.0 & \textbf{0.756} & 0.742 & \textbf{0.756} & 0.528 & 0.774 & & 35.0 & \textbf{0.705} & 0.708 & \textbf{0.738} & 0.456 & 0.761 \\
    \rowcolor{LightCyan}
    \cellcolor{white}\multirow{-6}{*}{\textbf{ViT-L/14}} & \textbf{\oursref} & & \textbf{89.8} & 0.741 & \textbf{0.744} & 0.750 & \textbf{0.530} & \textbf{0.776} & & \textbf{41.3} & 0.695 & \textbf{0.715} & 0.737 & \textbf{0.468} & \textbf{0.770} \\
    \bottomrule
  \end{tabular}
    }
\vspace{-0.2cm}
\end{table*}

%% file: sections/06_conclusion.tex
\section{Conclusion}
In this paper, we have presented \ours, a novel learnable metric aimed at improving the training and evaluation of captioning models. Leveraging a positive-augmented contrastive learning strategy in conjunction with a LoRA fine-tuning stage, \ours enhances the alignment between images and textual descriptions, proving effective in both evaluation and training phases. Our approach outperforms existing reference-based and reference-free metrics in terms of correlation with human judgment and sensitivity to object hallucinations, providing a promising pathway for advancing the quality and robustness of captioning models. Furthermore, experimental results demonstrate that incorporating \ours as a reward during the SCST fine-tuning phase significantly improves the quality of generated captions, mitigating issues like word repetition and hallucination. These findings underscore the potential of \ours to substantially enhance both the quality of generated captions and the accuracy of their evaluation.

%% file: sections/suppl.tex
\clearpage
\begin{appendices}
\section{}

\subsection{Out-of-domain Evaluation}
In Table~\ref{tab:zeroshot-cc3m}, we report additional out-of-domain results, and we evaluate the models on the CC3M dataset, which includes image-caption pairs sourced from web repositories. The results show consistency with those observed on the nocaps and VizWiz datasets, reported in the main paper. Notably, employing \ours as reward consistently enhances semantic richness while preserving fluidity during generation, as demonstrated by the higher CIDEr scores than those achieved by the model optimized via CLIP-S reward. This improvement is evident across both ViT-B/32 and ViT-L/14 backbones, further confirming the effectiveness of our training strategy and its generalization capabilities to out-of-domain datasets.

\begin{table*}[b]
  \centering
    \caption{Captioning results in terms of reference-based and reference-free evaluation metrics on CC3M validation set.}
  \label{tab:zeroshot-cc3m}
  \setlength{\tabcolsep}{.3em}
  \resizebox{0.7\linewidth}{!}{
  \begin{tabular}{lc c cccc>{\color{black}}cc}
    \toprule
    & & & \multicolumn{6}{c}{\textbf{CC3M}} \\
    \cmidrule{4-9}
    \textbf{Backbone} & \textbf{Reward} & & C & CLIP-S & PAC-S++ & RefCLIP-S & Polos & RefPAC-S++ \\
    \midrule
    & - & & 22.8 & 0.643 & 0.653 & 0.638 & 0.265 & 0.688 \\
    & CIDEr & & 27.9 & 0.655 & 0.663 & 0.657 & 0.253 & 0.698 \\
    \cmidrule{2-9}
    & CLIP-S & & 0.6 & \textbf{0.710} & 0.691 & 0.579 & 0.300 & 0.675 \\
    \rowcolor{LightCyan}
    \cellcolor{white}  & \textbf{\ours} & & \textbf{9.5} & 0.702 & \textbf{0.697} & \textbf{0.621} & \textbf{0.318} & \textbf{0.686} \\
    \cmidrule{2-9}
    & Ref-CLIP-S & & 21.1 & \textbf{0.679} & 0.678 & 0.660 & \textbf{0.303} & 0.702 \\
    \rowcolor{LightCyan}
    \cellcolor{white}\multirow{-6}{*}{\textbf{ViT-B/32}} & \textbf{\oursref} & & \textbf{24.6} & 0.676 & \textbf{0.686} & \textbf{0.661} & 0.301 & \textbf{0.709} \\
    \midrule
    & - & & 27.1 & 0.653 & 0.665 & 0.648 & 0.304 & 0.699 \\
    & CIDEr & & 34.8 & 0.665 & 0.672 & 0.676 & 0.273 & 0.713 \\
    \cmidrule{2-9}
    & CLIP-S & & 0.8 & \textbf{0.726} & \textbf{0.708} & 0.609 & 0.342 & 0.690 \\
    \rowcolor{LightCyan}
    \cellcolor{white}  & \textbf{\ours} & & \textbf{13.8} & 0.715 &  \textbf{0.708} & \textbf{0.639} & \textbf{0.359} & \textbf{0.698} \\
    \cmidrule{2-9}
    & Ref-CLIP-S & & 27.2 & \textbf{0.696} & 0.693 & 0.678 & 0.342 & 0.719 \\
    \rowcolor{LightCyan}
    \cellcolor{white}\multirow{-6}{*}{\textbf{ViT-L/14}} & \textbf{\oursref} & & \textbf{32.0} & 0.688 & \textbf{0.697} & \textbf{0.681} & \textbf{0.343} & \textbf{0.724} \\
    \bottomrule
  \end{tabular}
    }
\end{table*}
\begin{figure*}[b]
    \centering    \includegraphics[width=0.99\linewidth,height=0.3\linewidth]{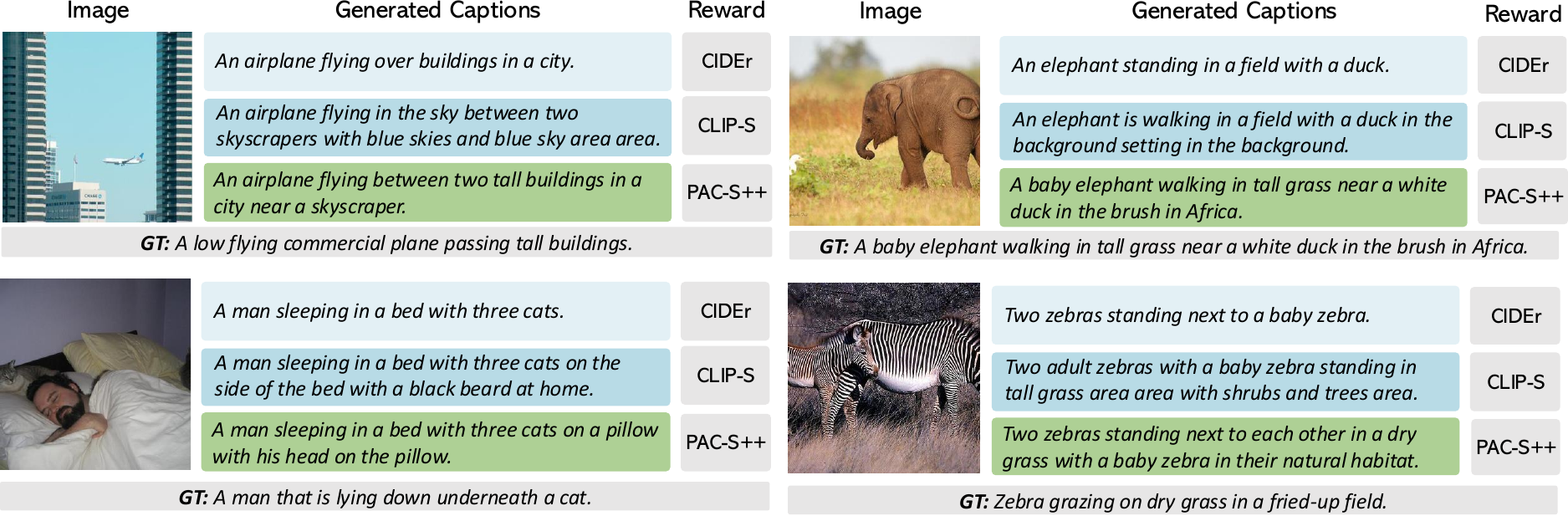}
    \caption{\rev{Sample failure cases of a captioning model fine-tuned with different metrics as reward.}}
    \label{fig:failures}
\end{figure*} 

\subsection{\rev{Limitations}}
\rev{While the proposed metric demonstrates strong performance across several benchmarks, it is important to acknowledge its limitations and examine failure cases that suggest directions for future improvement. Notably, our model inherits some constraints from the underlying CLIP architecture, as both the textual and visual encoders in our framework are based on CLIP backbones. Although we apply task-specific finetuning through a positive augmentation strategy, the core architectural limitations remain.}

\rev{On the textual side, the main limitation stems from the fixed token length (\ie~77 tokens), which can lead to truncation and loss of semantic information in longer textual descriptions. This might restricts the model ability to fully leverage context-rich or multi-sentence inputs. On the visual side, the use of a fixed input resolution can cause the loss of fine-grained details, especially in high-resolution or complex scenes. Moreover, since CLIP generates a global image embedding, the model may struggle with tasks requiring fine-grained recognition or explicit modeling of spatial relationships between objects.}

\tit{\rev{Failure Cases}}
\rev{To provide some insights for future research, we also include representative failure cases in Fig.~\ref{fig:failures}, when employing \ours as reward to fine-tune a captioning model. While our approach already shows a clear reduction in repetitive captions, as shown in Table~\ref{tab:coco}, some challenges such as hallucination and object miscounting still persist. These are known limitations across captioning systems and are difficult to fully eliminate -- even with strong reward signals like \ours. In particular, our model can misjudge spatial relationships in the absence of explicit depth or perspective information, such as describing a plane as ``between'' two buildings when it is actually flying behind them. It can also struggle with accurately counting objects, especially when they are occluded or only partially visible. Subjective attributes, like distinguishing between ``tall'' and ``short'' grass, often remain ambiguous even to human observers, further complicating the task. Additionally, the model may occasionally hallucinate plausible yet nonexistent background details. Although these errors are relatively rare and often subtle, they highlight ongoing challenges in the field and point toward meaningful directions for future improvement.}

\subsection{Additional Qualitative Results}
\begin{figure*}[t]
    \centering
    \includegraphics[width=0.99\linewidth]{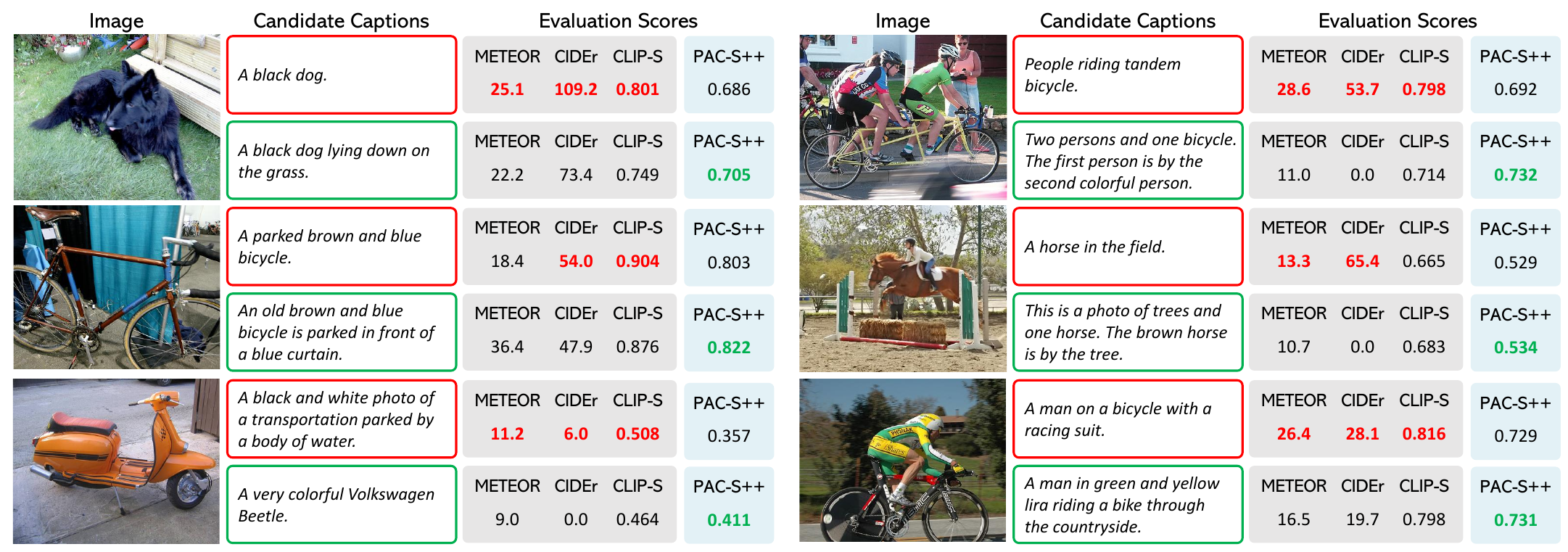}
    \caption{Evaluations of existing metrics for captioning evaluation in comparison to \ours on the Pascal-50S dataset. The candidate caption, preferred by humans and highlighted in green, is emphasized for reference.}
    \label{fig:pascal_qualitatives}
    \vspace{-0.2cm}
\end{figure*}

\begin{figure*}[t]
    \centering
    \includegraphics[width=0.99\linewidth]{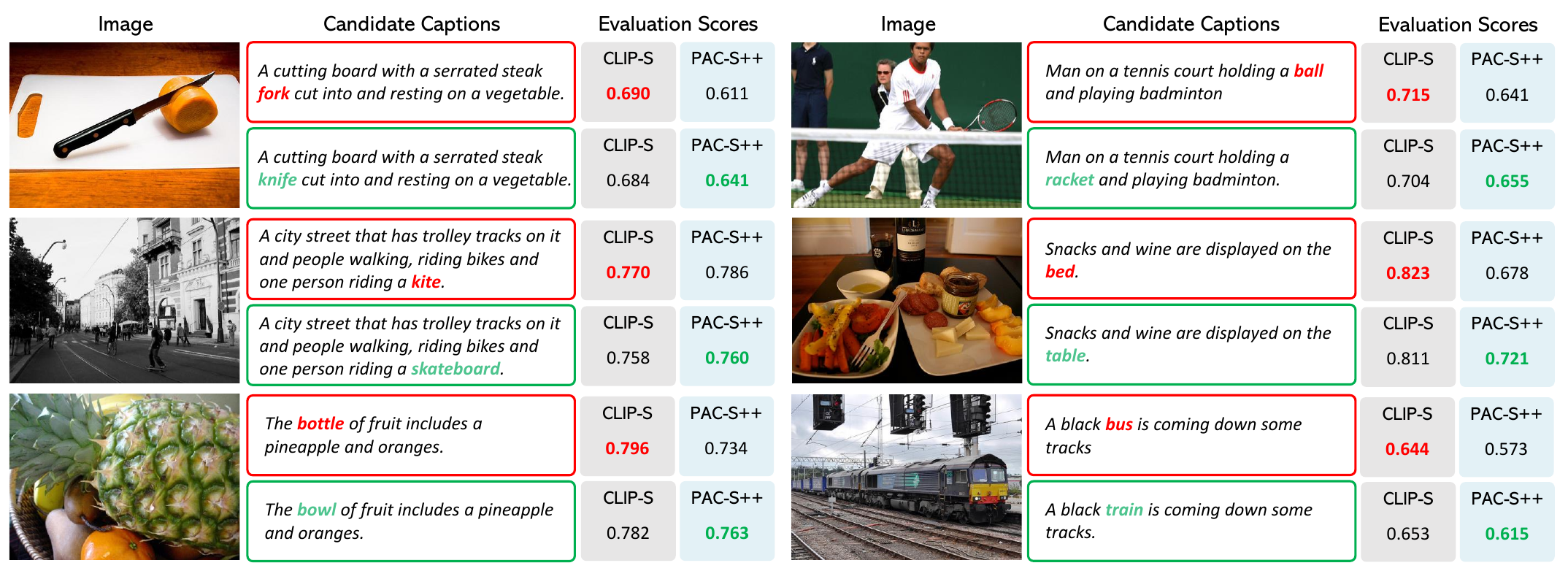}
    \caption{Sample images from the FOIL hallucination detection dataset and corresponding evaluation scores generated by our proposed metric in comparison with CLIP-S. Captions with hallucinated objects are highlighted in red.}
    \label{fig:foil_qualitatives}
    \vspace{-0.2cm}
\end{figure*}

In Fig.~\ref{fig:pascal_qualitatives}, we report qualitative results on the PASCAL50-S dataset, comparing \ours to well-known metrics. These qualitative results demonstrate that, in the majority of cases, \ours is more aligned with human judgment compared to other metrics.
Moreover, in Fig.~\ref{fig:foil_qualitatives}, we present sample results on the FOIL dataset. As shown in the figure, we compare the ability of \ours with CLIP-S in detecting hallucinated objects and demonstrate that \ours better correlates with human judgment and exhibits higher accuracy in correctly identifying hallucinated objects. 

\begin{figure*}[t]
    \centering
    \includegraphics[width=0.99\linewidth]{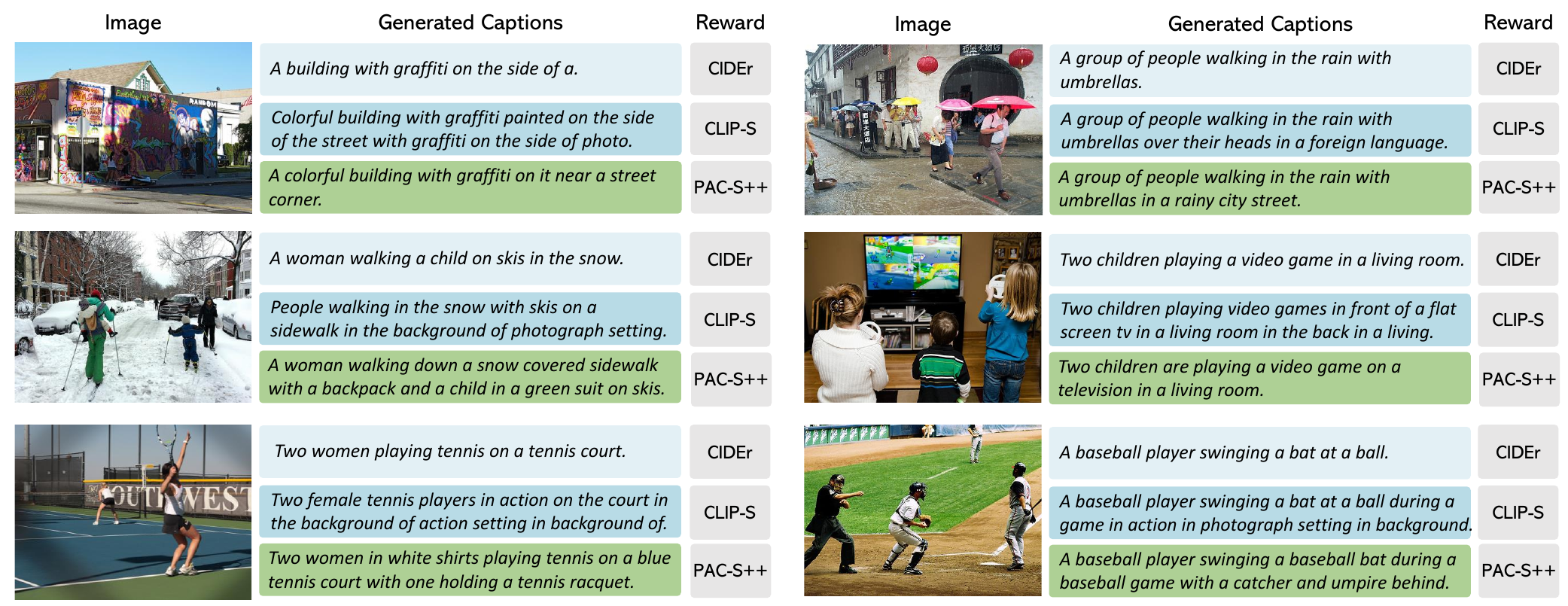}
    \caption{Additional qualitative image captioning results employing different metrics as reward.}
    \label{fig:reward_qualitatives_2}
    \vspace{-0.2cm}
\end{figure*}

Finally, in Fig.~\ref{fig:reward_qualitatives_2}, we present additional qualitative results obtained by using different types of rewards in the image captioning task. As it can be seen, employing \ours as reward leads to semantically richer captions without repetitions and grammatical errors, in contrast to generations observed with CLIP-S or CIDEr rewards.

\clearpage

\end{appendices}